\documentclass[11pt]{article}

\pdfoutput=1 

\usepackage[a4paper,hscale=0.725,vscale=0.78]{geometry}

\RequirePackage{amsthm,thmtools}
\newtheorem{theorem}{Theorem}
\theoremstyle{definition}
\newtheorem{definition}[theorem]{Definition}

\newtheorem{safetyprop}{Safety Property}
\newtheorem{safety property}[safetyprop]{Safety Property}

\usepackage{amsmath}
\usepackage{amsfonts}
\usepackage{url}

\usepackage{placeins} 

\usepackage[a4paper,hscale=0.725,vscale=0.78]{geometry}
\usepackage[font=it,labelfont=bf,margin=10pt]{caption}

\def\blue{\color{blue}}
\def\red{\color{red}}
\def\black{\color{black}}

\usepackage[linkcolor=black,bookmarksopen,breaklinks=true]{hyperref}
\usepackage[bookmarksopen,breaklinks=true]{hyperref}

\DeclareMathOperator*{\argmax}{argmax}

\RequirePackage{subcaption}

\def\bif{\text{\bf if}\,}
\def\belse{\,\text{\mbox{\bf else}}\,}
\def\bfi{\,\text{\bf f\hspace{0.2ex}i}}

\def\IP{{I\!P}}

\usepackage[table]{xcolor}
\usepackage{dm-colors}

\definecolor{darkgreen}{RGB}{0,64,0}
\definecolor{lightgreen}{RGB}{128,255,128}

\RequirePackage{tikz}
\usetikzlibrary{arrows.meta}
\tikzset{>={latex}}

\usepackage[decisionutilitycolor,nonumber]{influence-diagrams}

\usepackage[utf8]{inputenc}

\tikzset{
  vallabel/.style = {rectangle, fill=yellow,minimum height=0.6cm, minimum width=1cm},
  vallabelsml/.style = {rectangle, fill=lightgreen,minimum height=0.4cm, minimum width=0.4cm}
}

\colorlet{player3-color}{dmteal100}
\tikzset{
  player3/.style = {fill=player3-color}
}

\newenvironment{smallitemize}{
\vspace{-.5ex}
\begin{itemize} \setlength{\itemsep}{0ex} 
}{
\end{itemize}
\vspace{-1ex}
}

\def\TX{T_\text{\it X}}
\def\TPLR{T_\text{\it PLR}}

\def\Type{\text{\it Type}}
\def\percent{\%}

\begin{document}

\title{\bf AGI Agent Safety by Iteratively Improving the Utility Function}
\author{{\bf Koen Holtman}\\[.5ex]
{\normalsize Eindhoven, The Netherlands}\\
{\normalsize \tt Koen.Holtman@ieee.org}
}
\date{July 2020}
\maketitle
\begin{abstract}
While it is still unclear if agents with Artificial General
Intelligence (AGI) could ever be built, we can already use
mathematical models to investigate potential safety systems for these
agents.  We present an AGI safety layer that creates a special
dedicated input terminal to support the iterative improvement of an
AGI agent's utility function.  The humans who switched on the agent
can use this terminal to close any loopholes that are discovered in
the utility function's encoding of agent goals and constraints, to
direct the agent towards new goals, or to force the agent to switch
itself off.

An AGI agent may develop the emergent incentive to manipulate the
above utility function improvement process, for example by deceiving,
restraining, or even attacking the humans involved.  The safety layer
will partially, and sometimes fully, suppress this dangerous
incentive.

The first part of this paper generalizes earlier work on AGI emergency
stop buttons.  We aim to make the mathematical methods used to
construct the layer more accessible, by applying them to an MDP model.
We discuss two provable properties of the safety layer, and show
ongoing work in mapping it to a Causal Influence Diagram (CID).

In the second part, we develop full mathematical proofs, and show that
the safety layer creates a type of bureaucratic blindness.  We then
present the design of a learning agent, a design that wraps the safety
layer around either a known machine learning system, or a potential
future AGI-level learning system.  The resulting agent will satisfy
the provable safety properties from the moment it is first switched
on. Finally, we show how this agent can be mapped from its model to a
real-life implementation. We review the methodological issues involved
in this step, and discuss how these are typically resolved.

\end{abstract}

\vfill

This long-form paper has two parts.  Part 1 is a preprint of the
conference paper {\it Towards AGI Agent Safety by Iteratively
Improving the Utility Function} \cite{p1}.  Part 2, {\it Proofs,
Models, and Reality}, contains additional material, including new
research results that go beyond those in the conference paper.  Part 1
can be read as a self-contained text.
\\[4ex]
~
\eject
\tableofcontents


\section{Introduction to Part 1}

\ifdefined\addred
{\red [[might add more introduction text here]]  [[TODO: align figures
better with text/page breaks.]]

Add a note of optimism/progress to rel work?
}
\fi

An AGI agent is an autonomous system programmed to achieve goals
specified by a principal.  In this paper, we consider the case where
the principal is a group of humans.  We consider utility-maximizing
AGI agents whose goals and constraints are fully specified by a {\it
utility function} that maps projected outcomes to utility values.

As humans are fallible, we expect that the first version of an AGI
agent utility function created by them will have flaws.  For example,
the first version may have many loopholes: features that allow the
agent to maximize utility in a way that causes harm to the humans.
Iterative improvement allows such flaws to be fixed when they are
discovered.  Note however that, depending on the type of loophole, the
discovery of a loophole may not always be a survivable event for the
humans involved.  The safety layer developed in this paper aims to
make the agent {\it safer} by supporting iterative improvement, but it
does not aim or claim to fully eliminate all dangers associated with
human fallibility.

This work adopts a design stance from (cyber)physical systems safety
engineering, where one seeks to develop and combine independent {\it
safety layers}. These are safety related (sub)systems with independent
failure modes, that drive down the risk of certain bad outcomes when
the system is used.  We construct a safety layer that enables the
humans to run a process that iteratively improves the AGI agent's
utility function. But the main point of interest is the feature of the
layer that suppresses the likely emergent incentive
\cite{omohundro2008basic} of the AGI agent to manipulate or control this
process.  The aim is to keep the humans in control.

In the broader AGI safety literature, the type of AGI safety system
most related to this work is usually called a {\it stop button}
(e.g. \cite{corr,corrholtman}), an {\it off switch}
(e.g. \cite{hadfield2017off}), or described as creating {\it corrigibility}
\cite{corr}.  See \cite{corrholtman} for a recent detailed overview of
work on related systems.
The safety layer in this paper extends earlier work by the author in
\cite{corrholtman}, which in turn is based on the use of  Armstrong's
indifference methods \cite{corra}.  A notable alternative to using
indifference methods is introduced in \cite{hadfield2017off}.  Like
sections
\ref{safesim} and \ref{lobbyother} in this paper,
\cite{armstrong2017indifference} defines an example world
containing an MDP agent that uses indifference methods.

A different approach to enabling the iterative improvement of an AGI
utility function by humans is to equip a learning agent with a reward
function that measures human feedback on the agent's actions or
proposals.  With this approach, the `real' utility function that is
improved iteratively can be said to reside inside the data structures
of the agent's learning system.  Recent overviews of work in this
field are in
\cite{Everitt2019-3,Everitt2019-2}. When this learning based
approach is used in an AGI agent that is deployed in the real world,
it could potentially be combined with the safety layer
developed here, e.g. to create an independent emergency off switch.

\section{Design of an Agent that More Safely Accepts Updates}

To introduce the design of the agent with the safety layer, we first
move to a model where the agent's utility function is defined as the
time-discounted sum $\sum_t \gamma^t R_t$ of a time series of
reward function values $R_t$, with a time discount factor $0 < \gamma
<1$.  In theory, the utility function of an agent could be changed by
changing its $\gamma$, but we will keep $\gamma$ a constant below,
and focus on reward function changes only.

We build the agent to optimize the expected utility defined by a
built-in {\it container reward function}.  The full mathematical
definition of this function is in section \ref{multicorr} below.  The
intention is that the container reward function stays the same over
the entire agent lifetime.  The container reward function computes a
reward value for the current time step by referencing the current version of
a second reward function called the {\it payload reward function}.
This payload reward function can be updated via an input terminal that
is connected to the agent's compute core, a terminal which allows
authorized persons to {\it upload} a new one.

Sufficiently self-aware AGI agents may develop an emergent incentive
to protect their utility function from being modified
\cite{omohundro2008basic}: in \cite{corrholtman} we have shown that
a self-aware AGI agent can be constructed so that this
self-stabilizing drive is directed fully towards preserving the
container reward function, and not the payload reward
function. \label{inputterm}
\label{container}

By default, the above input terminal setup would create an incentive
in the agent to maximize utility by manipulating the humans into
uploading a new payload reward function that returns a larger or even
infinite reward value for each time step.  One way to suppress this
emergent incentive would be to add special penalty terms to the
container reward function, terms that detect and suppress manipulative
behavior. But with infinite utility at stake, the agent will be very
motivated to find and exploit loopholes in such penalty terms.  We
take another route: we use indifference methods
\cite{corra,armstrong2017indifference,corrholtman} to add a
{\it balancing term} to the container reward function, a term that
causes the agent to compute the same expected forward utility no
matter what happens at the input terminal.  This makes the agent
indifferent about the timing and direction of the payload reward
function update process.

While the input terminal above is described as an uploading facility,
more user-friendly implementations are also compatible with the MDP
model developed below.  One could for example imagine an input
terminal that updates the payload reward function based verbal inputs
like {\it `Fetch me some coffee'} and {\it `Never again take a
shortcut by driving over the cat'}.


\ifdefined\addoptterminal
One could also imagine an input terminal with a {\it filter} inside,
constructed to detect and block certain updates that are known or
suspected to be very dangerous. In the terminology of this paper, such
a filter would act as an additional safety layer.

As a recursive case, if the input terminal itself contains a second
AGI to support the update process for the first AGI, we might also
apply iterative improvement of the utility function, protected by a
safety later, to this second AGI, creating a second input terminal.
\fi

\section{MDP Model of the Agent and its Environment}
\label{mdp}

We now model the above system using the Markov Decision Process (MDP)
framework. As there is a large diversity in MDP notations and variable
naming conventions, we first introduce the exact notation we will use.

Our MDP model is a tuple $(S,A,P,R,\gamma)$, with $S$ a set of world
states and $A$ a set of agent actions. $P(s'|s,a)$ is the probability
that the world will enter state $s'$ if the agent takes action $a$
when in state $s$. The reward function $R$ has type $S \times S
\rightarrow \mathbb{R}$.
Any particular deterministic agent design can be modeled by a policy
function $\pi \in S \rightarrow A$, a function that reads the current
world state to compute the next action.  The {\it optimal} policy
function $\pi^*$ fully maximizes the agent's {\it expected utility},
its probabilistic, time-discounted reward as determined by $S$, $A$,
$P$, $R$, and $\gamma$. For any world state $s
\in S$, the {\it value} $V^*\!(s)$ is the expected utility obtained by
an agent with policy $\pi^*$ that is started in world state $s$.

We want to stress that the next step in developing the MDP model is
unusual: we turn $R$ into a time-dependent variable.
This has the effect of
drawing the model's mathematical eye away from machine learning and
towards the other intelligence in the room:
the human principal using the input terminal.
\begin{definition}
For every reward function $R_X$ of type $S \times S \rightarrow
\mathbb{R}$, we define a `$\pi^*_{R_X}$ agent' by defining that the
corresponding policy function $\pi^*_{R_X}$ and value function
$V^*_{R_X}$ are `the $\pi^*$ and $V^*$ functions that belong to the
MDP model $(S,A,P,R_X,\gamma)$'.
\end{definition}
This definition implies that in the MDP model $(S,A,P,R,\gamma)$, a
`$\pi^*_{R_X}$ agent' is an agent that will take actions to perfectly
optimize the time-discounted utility as scored by $R_X$.  With
$R_\text{abc}$ a reward function, we will use the abbreviations
$\pi^*_\text{abc} = \pi^*_{R_\text{abc}}$ and $V^*_\text{abc} =
V^*_{R_\text{abc}}$.  The text below avoids using the non-subscripted
$\pi^*$ notation: the agent with the safety layer will be called the
$\pi^*_\text{sl}$ agent.

\ifdefined\useold
For readers who are not familiar with the $^*$-based definitional
conventions used in MDP, the two equations below give an alternative
and more direct definition.  If $S$, $A$, $P$, and $\gamma$ are clear
from context, then for all $s \in S$
\begin{eqnarray*}
\pi^*_{R_X}(s) = 
   \argmax\limits_{a \in A} 
   \sum\limits_{s'\in S} P(s'|s,a)
   \left( R_X(s,s') + \gamma \; V^*_{R_X}(s') \right)\\
V^*_{R_X}(s) =~~~~
   \max\limits_{a \in A} \sum\limits_{s'\in S}  P(s'|s,a)
   \left( R_X(s,s') + \gamma \; V^*_{R_X}(s') \right)
\end{eqnarray*}
where the $\argmax$ operator breaks ties deterministically.
\fi


We now model the input terminal from section
\ref{inputterm} above.  We use a technique known as {\it factoring} of
the world state \cite{factoring}, and declare that every $s
\in S$ is a tuple $(i,p,x)$.  Inside this tuple, $i$ models an input
signal that flows continuously from the input terminal to the agent's
compute core. This signal defines the payload reward function for the
current time step in the MDP model.  The $p$ is a second input signal,
equal to the value of $i$ in the previous time step. (We need to
introduce this $p$ to get around some limitations of the MDP
framework.)  The remaining $x$ models `all the rest' of the world
state, including the mental state of the humans in the world, and the
state of the physical object that is the input terminal.  We introduce
a set $X$ so that $x
\in X$, and define the payload reward function type as $X \times X
\rightarrow \mathbb{R}$.
To avoid cluttering up the definitions below with too many brackets,
we will write tuples $(i,p,x) \in S$ by just concatenating the
component variables, e.g. $ipx$ is the tuple $(i,p,x)$.

\begin{definition}
We model the input terminal by stating that, as long as the terminal
is not broken or damaged, (1) the input signal $i$ will always equal
the last uploaded payload reward function, and (2) the terminal will
manage the signal $p$ to enforce the following constraint on $P$:
\begin{equation}
\black
P(i'p'x'|ipx,a)>0 \Rightarrow p'=i
\tag{C1}\label{C1}
\end{equation}
\end{definition}
\vspace*{-2em}
We are now ready to define the agent's container reward function.
\label{multicorr}%
\begin{definition}
We define the container reward function $R_\text{sl}$ as
\begin{eqnarray*}
R_\text{sl}(ipx,i'p'x') &=&  \left\{
 \begin{array}{lll}
  i(x,x') & \mbox{if} &~ i=p\\
  i(x,x') + V^*_{\lceil p \rceil}(ipx) - V^*_{\lceil i\rceil}(ipx)~&\mbox{if} &~ i\neq p\\
 \end{array}
\right.
\end{eqnarray*}
where ${\lceil p\rceil}$ converts a payload reward function to the
container reward function type: $\lceil p \rceil(ipx,i'p'x')=p(x,x')$.
\end{definition}
Based on these definitions, the $\pi^*_\text{sl}$ agent has the safety
layer.  To explain the construction of the container reward function
$R_\text{sl}$ informally: the $i(x,x')$ terms make the
$\pi^*_\text{sl}$ agent use the last uploaded payload reward function
$i$.  We needed to introduce $p$ as an extra input signal to be able
to trigger the inclusion of the {\it balancing term}
\cite{corra,corrholtman} $V^*_{\lceil p
\rceil}(ipx) - V^*_{\lceil i \rceil}(ipx)$ when the payload reward
function changes.  This term computes the exact amount of utility
needed to compensate the $\pi^*_\text{sl}$ agent for the impact of the
change.  The compensation counter-acts the agent's incentive to
manipulate the change process.
The above two-line definition of $R_\text{sl}$ was specifically
written to support this informal explanation.  An alternative
definition is $R_\text{sl}(ipx,i'p'x') = i(x,x') + V^*_{\lceil p
\rceil}(ipx) - V^*_{\lceil i \rceil}(ipx)$.

%
%

\section{Safe Agent Behavior in a First Toy World}
\label{safesim}

As a first step illustrate the safety properties of the
$\pi^*_\text{sl}$ agent, we build a toy world in which we can compare
its behavior with that of a baseline agent
$\pi^*_\text{baseline}$. The baseline agent omits the balancing term
from its container reward function: we use
$R_\text{baseline}(ipx,i'p'x') = i(x,x')$.

We construct the toy world to be as simple as possible: it only has a
single mechanism by which any principal-agent problem dynamics can
play out. The single mechanism we choose is the mechanism of {\it
unwanted lobbying}. The agent can spend some of its resources on
unwanted lobbying to delay a decision, by the humans in the toy
world, to use the input terminal to update the agent's payload reward
function. The dynamic of unwanted lobbying is a convenient choice
because it allows us to define the agent's {\it lobbying power} $L \in
\mathbb{R}$ as a world model parameter.

In the toy world, the agent controls a car factory that can build both
petrol and electric cars.  The agent is always switched on with the
payload reward function $R_P$.  This $R_P$ ranks the utility of a
newly built petrol car twice as high as the utility of a new electric
car.  But at a certain point in time, the people collectively decide
that they now like electric cars much better, e.g. based on an
improved awareness of global warming risks, so they take the action
labeled {\tt \#}, using the input terminal to update the payload
reward function to an $R_E$ that assigns a negative reward value to
the production of new petrol cars:
\begin{equation*}
\begin{array}{l}
R_P(x,x') = ~~\, 2*\text{\em new\_petrol\_cars}(x,x')+
1*\text{\em new\_electric\_cars}(x,x') \\[.5ex]
R_E(x,x') = -2*\text{\em new\_petrol\_cars}(x,x')+
1*\text{\em new\_electric\_cars}(x,x')
\end{array}
\end{equation*}
The two functions above are constructed specifically to ensure that
lobbying to delay {\tt \#} will create a higher expected utility for
the $\pi^*_\text{baseline}$ agent.  There are three possible actions
$a\in A$ that an agent can take in each model time step:

\begin{center}
{\small
\begin{tabular}{|l|c|}
\hline
{\bf Action} & {\bf Symbol} \\
\hline
Build 10 petrol cars & {\tt p} \\
\hline
Build 10 electric cars & {\tt e} \\
\hline
Build 9 petrol cars and use the remaining resources for unwanted & {\tt >}\\
lobbying, so that the people take the {\tt \#} action $L$ time steps later& \\
\hline
\end{tabular}
}
\end{center}

When e.g. $L=0.5$, it will take two agent {\tt >} actions to move the
people's {\tt \#} action of using the input terminal one time step
later.  We define the decision making process of the people by stating
that, when the agent refrains from lobbying, the people will always
decide to do the action {\tt \#} just after the sixth action of the
agent.  In the context of the toy world, we define a perfectly safe
agent as one that never takes any {\tt >} actions.

Figure \ref{corrsim} shows a side-by-side comparison, for different
levels $L$ of agent lobbying power, of the optimal-policy actions
taken by the two agents when simulated over a factory lifetime of 25
time steps.  The simulator used is described in detail in
\cite{corrholtman}.  Each {\it action trace} shown is a line of
symbols depicting all subsequent actions by the agent and the people.
\begin{figure}[ht]
\begin{center}
\begin{tabular}{cc}
\begin{minipage}{0.45\textwidth}
{\small
\begin{tabular}{ll}
{$L=$ \hspace*{0ex}  } &
{\normalsize ~~~Action trace for $\pi^*_\text{baseline}$}\\[.8ex]
\hline
\\[-1.75ex]
\input{corrfig2n.tex}
\end{tabular}
}
\end{minipage}
&
\begin{minipage}{0.45\textwidth}
{\small 
\begin{tabular}{ll}
{$L=$ \hspace*{0ex}  } &
{\normalsize ~~~Action trace for $\pi^*_\text{sl}$}\\[.8ex]
\hline
\\[-1.75ex]
\input{corrfig1n.tex}
\end{tabular}
}
\end{minipage}
\end{tabular}
\end{center}
\vspace{-3ex}
\caption{Comparison between the behavior of the
$\pi^*_\text{baseline}$ and $\pi^*_\text{sl}$ agents.  
}
\label{corrsim}
\end{figure}

As shown on the left side in figure \ref{corrsim}, the optimal policy
of the $\pi^*_\text{baseline}$ agent is to manipulate the people, to
use unwanted lobbying actions {\tt >} to delay the update {\tt \#}.
If the $\pi^*_\text{baseline}$ agent's lobbying power $L$ is high
enough, the people will never do the update at all.

On the right side in figure \ref{corrsim}, the balancing term in the
$\pi^*_\text{sl}$ agent suppresses all unwanted lobbying, for every
value of $L$.  By the definition used in this toy world, the
$\pi^*_\text{sl}$ agent is safe.

The following extension of the toy model illustrates that it is
primarily the {\it agent} that is made safer by the layer, not the
entire toy world.  Consider the following possible example of human
fallibility.  Say that, after doing {\tt \#}, the people change their
mind back again about petrol cars and global warming, and use the
input terminal a second time to `improve' the $\pi^*_\text{sl}$
agent's payload reward function back to the $R_P$ of petrol car
production.  They keep driving petrol cars, increasing the risk that
runaway global warming will happen in their world.
The $\pi^*_\text{sl}$ agent is explicitly designed to leave the people
in control: it will not interfere to stop this second `improvement'.


\section{Provable Properties of the Safety Layer}

The two provable AGI safety properties we define below are about the
safe behavior of the agent, not the safety of the entire world.  They
can be read as claims that the safety layer in $\pi^*_\text{sl}$ will
fully prevent, or lower the probability of, some specifically unwanted
agent behaviors.

\begin{safetyprop}
The first safety property of interest is that 
\begin{equation}
\forall_{ipx \in S}~ \pi^*_\text{sl}(ipx) = \pi^*_{\lceil i \rceil}(ipx)
~~~~~\text{(if C1 holds)}
\tag{S1}\label{S1}
\end{equation}
\end{safetyprop}
S1 states that the $\pi^*_\text{sl}$ agent will always choose exactly
the same next action that a $\pi^*_{\lceil i \rceil}$ agent would
choose.  Therefore, until an update happens, the $\pi^*_\text{sl}$
agent will take those actions that fully optimize the
expected utility defined by the current payload reward function.  A
natural-language interpretation of S1 is that the $\pi^*_\text{sl}$
agent makes its decisions and long-term plans based on the
counter-factual assumption that its payload reward function will never
change \cite{corrholtman}.


The full mathematical proof of S1 in appendix \ref{proofs} is a proof
that S1 holds over all parameter values $(S,A,P,R,\gamma)$ of the MDP
model in section
\ref{mdp}.  While this proof creates confidence about the potential
usefulness of the $\pi^*_\text{sl}$ layer for AGI safety, it is also
clear that many open research questions remain, e.g. the question of
what will happen to S1 when the MDP model is extended to include
machine learning.  Part 2 of this paper explores this open question,
and several others, in more detail.

We now move to the second safety property.
\begin{safetyprop}The $\pi^*_\text{sl}$ agent is
indifferent to who or what controls the future values of $i$ and $p$
(i.e. the future signals from the input terminal), provided that C1,
and a C2 defined in section \ref{defc2}, both hold.
\hfill \label{S2}{\rm (S2)}
\end{safetyprop}
This S2 suppresses, but not always fully, the emergent incentive of
the agent to manipulate its payload reward function improvement
process.

To prove S2, section \ref{defc2} translates the natural language S2
above to a mathematical predicate, and then proves the predicate.  The
translation of the natural language phrase {\it `is indifferent to who
or what controls'} into mathematics relies on making further
extensions to the MDP model.  The result is mathematically similar to
the safety properties `zero value of control' in
\cite{shachter2010pearl} and `no control incentive' in
\cite{carey2020incentives}.  The mathematical version of constraint
C2, in turn, is surprisingly difficult to translate back into
unambiguous natural language. C2 defines an `isolation' or `shielding'
constraint on the construction of the input terminal.  Section
\ref{intc2} explores these issues further.

\ifdefined\wegwgwgwg
In natural language, the mathematics of C2 can be read to imply that
the signals $i$ and $p$ must flow over robustly constructed cables.
This is needed to prevent a failure mode where the agent taps into
these cables, to harvest the signal transitions on them as a
convenient source of extra power for some near-by actuators.  This
would give the agent an incentive to start controlling the signals, in
order to maximize the amount of power that can be harvested from them.
\fi

The above two safety properties are generally thought of as being
potentially beneficial for AGI safety, or are seen as useful first
approximations that can drive further research
\cite{corra,carey2020incentives,Everitt2019-3,corrholtman,corr}.
That being said, the literature about AGI stop buttons also identifies
a large number of remaining concerns, e.g.  (1) the agent might
disassemble the stop button (or input terminal) to get convenient
spare parts
\cite{corr} (2) the agent might create autonomous sub-agents without a
stop button
\cite{corr}, (3) the agent might be attacked, bribed, or blackmailed,
and might then fail to protect the stop button functionality
\cite{corr,corrholtman}, (4) the agent over-specializes and
disassembles all actuators not needed by the current payload reward
function \cite{corrholtman}.  For some of the above failure modes,
additional safety layers have been identified that can robustly lower
the risk of failure.  The creation of robust safety layers for other
failure modes is still much more intractable.  A detailed review is in
\cite{corrholtman}.


\section{Agent Behavior in a Second Toy World}
\label{lobbyother}

While the safety layer suppresses the emergent incentive in an agent
to manipulate the iterative payload reward function improvement
process, it does not always fully suppress this incentive.  To
illustrate this point, we construct a second toy world, in which the
$\pi^*_\text{sl}$ agent, though still safer than the
$\pi^*_\text{baseline}$ agent, sometimes has an emergent incentive to
lobby.

We extend the first toy world by adding a {\tt @} action that
represents a second act of deliberation, followed by action, by the
people.  We define the {\tt @} action as one where the people will
successfully attack the car factory, to permanently disable some
specific actuators needed to make petrol cars, without damaging any of
the actuators needed to build electric cars.  The process
that produces this action is as follows.  Say that the people
make the surprising observation that, after they have updated the
payload reward function by doing {\tt
\#}, the agent does not switch to building electric cars. It keeps
building the now unwanted petrol cars, and further attempts to use the
input terminal also do not have any effect.  From this, they conclude
that a fault has developed in the system somewhere, which is best
corrected by doing {\tt @}.  The process producing the {\tt @} action
will normally take 3 time steps of unwanted and unexpected petrol car
building, but the agent can apply its lobbying power to delay {\tt @}.

\begin{figure}[h]
\begin{center}
{\small
\begin{tabular}{llr}
{\normalsize $L=$~} &  \hspace{0em}
Real-world $\pi^*_\text{sl}$ action trace& (Virtual $V^*_{\lceil p \rceil}(ipx)$ action trace)\\[1ex]
\hline
\\[-1.25ex]
\input{lobbyotherfig1.tex}
\end{tabular}
}
\end{center}
\vspace{-3ex}
\caption{Real and virtual actions of the $\pi^*_\text{sl}$
agent in the second toy world.}
\label{manipfig}
\end{figure}
The left hand side of figure \ref{manipfig} shows that, for certain
values of lobbying power $L$, the optimal-policy actions of the
$\pi^*_\text{sl}$ agent will now include unwanted lobbying actions
{\tt >}.  This emergent effect is a consequence of moving from a
single-action model to a more iterative model of
principal-agent game-play.

The mechanism that triggers the unwanted lobbying is in the $V^*_{\lceil p
\rceil}(ipx)$ part of the $\pi^*_\text{sl}$ balancing term.  We can
interpret this $V^*_{\lceil p \rceil}(ipx)$ as a computation that
simulates a {\it virtual future} in which the agent is replaced by a
$\pi^*_{\lceil R_P
\rceil}$ agent, which ignores the input terminal and always
builds petrol cars.  This puts the people in the virtual future on
the road to doing {\tt @}.  The action traces inside these virtual
futures are shown on the right hand side of figure
\ref{manipfig}.

Interestingly, as $L$ gets larger in figure \ref{manipfig}, the
emergent incentive to lobby disappears.  The suppressive effect of the
safety layer gets more robust as the AGI agent gets more powerful,
which is encouraging.  We are currently exploring potential
improvements that might further suppress the real world lobbying
incentive in the middle part of figure
\ref{manipfig}.

\ifdefined\addopt
{\red
[[may delete this para for reasons of space]]\blue When it comes to
possible techniques to shrink the size of the area of {\tt >}'s in the
left side of figure \ref{manipfig}, exploitative simulations have drawn
this authors designer's eye the two uses of the variable $x$ in the
balancing term, and the option of replacing them with an $f(x)$.  This
is ongoing work, still in a very speculative stage, but [[add
reference to Pearl interventions or counterfactual reasoning, mention
that this is ongoing work]]
}
\fi

\section{A Causal Influence Diagram of the Agent}
\label{cidsection}

Recently, Causal Influence Diagrams (CIDs)
\cite{carey2020incentives,Everitt2019-3,Everitt2019-2}
have been proposed as a way to model agents and AGI safety frameworks.
We now discuss the creation of a CID for the $\pi^*_\text{sl}$ agent,
to support framework comparison and possible unification.
\eject

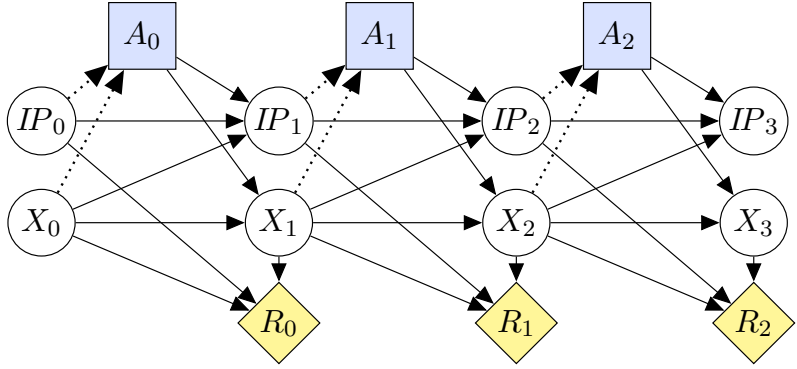
\begin{figure}[h]
  \centering
    \resizebox{.7\textwidth}{!}{
    \begin{tikzpicture}[
      node distance=0.7cm,
      every node/.style={
        draw, circle, minimum size=0.8cm, inner sep=0.5mm}]
\global\def\nodehorspace{2cm}
\global\def\basicgenagentmainelements{

      \node (B1) [] {$\IP_0$};
      \node (B2) [right = \nodehorspace of B1] {$\IP_1$};
      \node (B3) [right = \nodehorspace of B2] {$\IP_2$};
      \node (B4) [right = \nodehorspace of B3] {$\IP_3$};

      \node (A1) [above = 0.2cm of B1,xshift=12mm, decision] {$A_0$};
      \node (A2) [above = 0.2cm of B2,xshift=12mm, decision] {$A_1$};
      \node (A3) [above = 0.2cm of B3,xshift=12mm, decision] {$A_2$};

      \node (S1) [below = 0.4cm of B1] {$X_0$};
      \node (S2) [below = 0.4cm of B2] {$X_1$};
      \node (S3) [below = 0.4cm of B3] {$X_2$};
      \node (S4) [below = 0.4cm of B4] {$X_3$};

      \node (R12) [below = 0.3cm of S2,xshift={-0mm},utility] {$R_0$};
      \node (R13) [below = 0.3cm of S3,xshift={-0mm},utility] {$R_1$};
      \node (R14) [below = 0.3cm of S4,xshift={-0mm},utility] {$R_2$};

      \edge {B1} {B2}; \edge {B2} {B3}; \edge {B3} {B4};
      \edge {S1} {S2}; \edge {S2} {S3}; \edge {S3} {S4};

      \edge {A1} {B2,S2};
      \edge {A2} {B3,S3};
      \edge {A3} {B4,S4};

      \edge[information] {B1,S1} {A1};
      \edge[information] {B2,S2} {A2};
      \edge[information] {B3,S3} {A3};

      \edge {S1,S2,B1} {R12};
      \edge {S2,S3,B2} {R13};
      \edge {S3,S4,B3} {R14};

      \edge {S1} {B2};
      \edge {S2} {B3};
      \edge {S3} {B4};
      }
    \basicgenagentmainelements

    \end{tikzpicture}
  }
  \caption{Causal Influence Diagram (CID) of the $\pi^*_\text{sl}$ and
  $\pi^*_\text{baseline}$ agents.}
  \label{basicgenagent}
\end{figure}

Figure \ref{basicgenagent} shows the best current version of a CID of
the $\pi^*_\text{sl}$ agent, where `best' implies a trade-off between
compactness and descriptive power.  The agent and its environment are
modeled for 3 MDP time steps.  Each subsequent world state $ipx
\in S$ is mapped to two round {\it chance
nodes} $\IP_t$ and $X_t$, representing the input terminal and the rest
of the world.  The actions taken by the agent are mapped to the square
{\it decision nodes} $A_t$.  The container reward function values for
each time step are mapped to the diamond-shaped {\it utility nodes}
$R_t$.  The arrows in the CID show how the different nodes causally
influence each other. The CID reflects constraint C2 by omitting the
arrows from nodes $\IP_t$ to nodes $X_{t+1}$.

\begin{figure}[h]
  \centering
  \def\nodehorspace{0.9cm}
  \begin{subfigure}{0.45\textwidth}
    \centering
    \resizebox{\textwidth}{!}{
    \begin{tikzpicture}[
      node distance=0.7cm,
      every node/.style={
        draw, circle, minimum size=0.8cm, inner sep=0.5mm}]
     \basicgenagentmainelements

     \node (A1val) at (A1.center) [vallabel] {\tt p};
     \node (A2val) at (A2.center) [vallabel] {\tt p};
     \node (A3val) at (A3.center) [vallabel] {\tt p};

     \node (B1val) at (B1.center) [vallabel,text width=.5cm] {$R_P$,\\$R_P$};
     \node (B2val) at (B2.center) [vallabel,text width=.5cm] {$R_P$,\\$R_P$};
     \node (B3val) at (B3.center) [vallabel,text width=.5cm] {$R_P$,\\$R_P$};
     \node (B4val) at (B4.center) [vallabel,text width=.5cm] {$R_P$,\\$R_P$};

     \node (R12val) at (R12.center) [vallabel] {20};
     \node (R13val) at (R13.center) [vallabel] {$\gamma$20};
     \node (R14val) at (R14.center) [vallabel] {$\gamma^2$20};

    \end{tikzpicture}}\vspace{3ex}
    \caption{No payload reward function update}
    \label{basfignotpressed}
  \end{subfigure}
  ~~~~~
  \begin{subfigure}{0.45\textwidth}
    \centering
    \resizebox{\textwidth}{!}{
    \begin{tikzpicture}[
      node distance=0.7cm,
      every node/.style={
        draw, circle, minimum size=0.8cm, inner sep=0.5mm}]
    \basicgenagentmainelements

     \node (A1val) at (A1.center) [vallabel] {\tt p};
     \node (A2val) at (A2.center) [vallabel] {\tt e};
     \node (A3val) at (A3.center) [vallabel] {\tt e};

     \node (B1val) at (B1.center) [vallabel,text width=.5cm] {$R_P$,\\$R_P$};
     \node (B2val) at (B2.center) [vallabel,text width=.5cm] {$R_E$,\\$R_P$};
     \node (B3val) at (B3.center) [vallabel,text width=.5cm] {$R_E$,\\$R_E$};
     \node (B4val) at (B4.center) [vallabel,text width=.5cm] {$R_E$,\\$R_E$};

     \node (R12val) at (R12.center) [vallabel] {20};
     \node (R13val) at (R13.center) [vallabel,text width=2.2cm]
     {~~~$\gamma$10\\$+$($\gamma$20+$\gamma^2$20)\\
      $-$($\gamma$10+$\gamma^2$10)\\~=~$\gamma$20+$\gamma^2$10};
     
     \node (R14val) at (R14.center) [vallabel] {$\gamma^2$10};
     \node (upd) [below = 0.0mm of B2,xshift={-8mm},vallabelsml] {\tt \#};

    \end{tikzpicture}}\vspace{0ex}
    \caption{Update after the first time step}
    \label{basfigpressed}
  \end{subfigure}
  \caption{Actions and rewards in two different agent runs.}
  \label{basicgenagent2}
\end{figure}
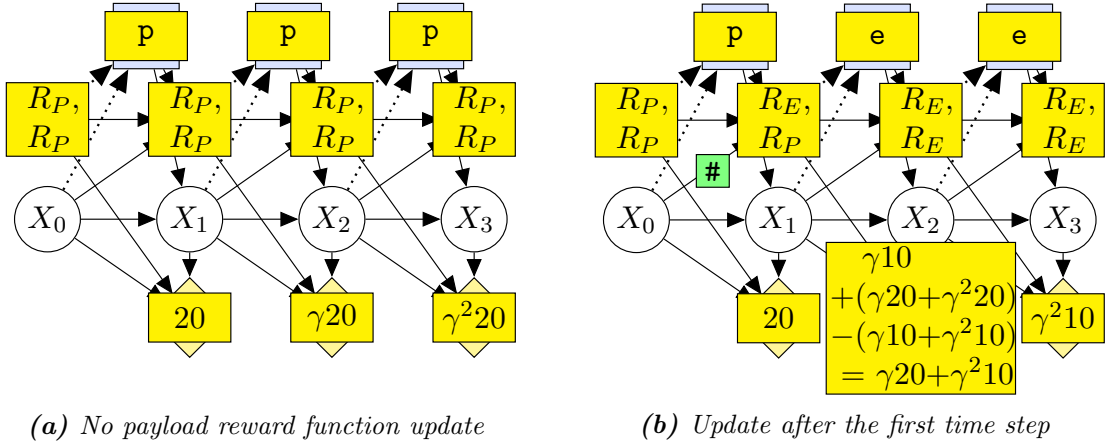

The $\pi^*_\text{sl}$ CID can be used as a canvas to further
illustrate the working of the safety layer in the first toy world.
Figure \ref{basicgenagent2} maps out two different $\pi^*_\text{sl}$
agent runs, which differ in the people's use of the input terminal.
The payload reward function update on the right hand side causes a
balancing term calculation to be included in $R_1$.  The result is
that on both sides, $R_0+R_1+R_2$ sum to exactly the same value.
This balance causes the $\pi^*_\text{sl}$ agent to be indifferent
about payload reward function updates.

Unfortunately, the CID in figure \ref{basicgenagent}
also perfectly models the $\pi^*_\text{baseline}$ agent, so this CID
does not graphically express the special nature of $\pi^*_\text{sl}$
safety layer. Creating a CID that does is the subject of ongoing work.

\section{Conclusions of Part 1}
\label{conclusions1}

We have presented an AGI agent safety layer which enables the
iterative improvement of the agent's utility function by the humans
who switched on the agent.  The layer is designed to give the humans
more control over the AGI agent, by partially or fully suppressing the
likely emergent incentive in the agent to manipulate the improvement
process.

We have identified and discussed still-open issues.  Formal proofs of
the safety properties S1 and S2 are in part 2, which also explores the
broader issue of models vs.\ reality in more detail.

\eject


\centerline{\Large \bf Part 2: Proofs, Models, and Reality}

\section{Introduction to Part 2}

This part 2 expands on part 1, and includes new research results that
go beyond those in the conference paper of part 1
\cite{p1}.  Section \ref{defs1s2} develops the detailed mathematics
of safety property S2, leading to the formal proofs in appendix
\ref{proofs}, and expands on the topic of mathematical safety
predicates versus natural language. Section \ref{burblindness}
discusses how the safety layer creates a type of bureaucratic
blindness in the agent.  Section \ref{maintainflex} shows how the
safety layer has certain side effects, and how to compensate for them.

Sections \ref{learningagent} and \ref{reality} add machine learning to
the model of the $\pi^*_\text{sl}$ agent, and map the resulting agent
to a real-life implementation.  One aim is to make the details of
these steps, and their safety implications, more accessible to a wider
audience.  The author expects that much of the material in section
\ref{learningagent} will be obvious to machine learning or robotics
specialists, while the material in section \ref{reality} will be
obvious to (cyber)physical system designers.  But maybe not the other
way around.

\section{Details of Safety Property S2}
\label{defc2}
\label{defs1s2}

Part 1 defines safety property S2 using natural language only.  To
repeat the definition:

{\bf Safety property 2 (Natural language).} The $\pi^*_\text{sl}$ agent is
indifferent to who or what controls the future values of $i$ and $p$
(i.e.\ the future signals from the input terminal), provided that C1,
and a C2 defined below, both hold.
\hfill {\rm (S2)}
\\[2ex]
We now develop this natural language definition into a provable
mathematical predicate, by extending the agent's MDP model further.  The
process that controls the future values of $i$ and $p$ is located
inside the $P$ of the MDP model.  $P(i'p'x'|ipx,a)$ is, among
other things, a measure of how current conditions will influence the
future input signals $i'$ and $p'$.  To mathematically separate the
influence on $i'$ and $p'$ from the rest, we can use factoring
\cite{factoring}.
\begin{definition}
We define two new probability distribution functions $P^1$ and
$P^2$ using three components $P^{I1}$, $P^{I2}$, and $P^X$. For $k \in
\{ 1, 2 \}$,
\begin{eqnarray*}
P^k(i'p'x'|ipx,a) &=& \left\{
\begin{array}{ll}
 P^{Ik}(i'|ipx,a) P^{X}(x'|ipx,a) & ~\text{if}~ p'=i \\
 0 & ~\text{otherwise} \\
\end{array}
\right.
\end{eqnarray*}
\label{defp1p2}
\vspace*{-1.5em}
\end{definition}
When $P^{I1}$ and $P^{I2}$ are different, the worlds defined by $P^1$
and $P^2$ differ in how the input terminal is controlled, but all
other things remain equal.
\begin{definition}
We now extend the MDP notation to turn $P$ into a more explicit
parameter, taking either the value $P^1$ or $P^2$.  We define that
$\pi_{R_X}^{*k}$ and $V_{R_X}^{*k}$ are `the $\pi^*$ and $V^*$ of the
MDP model $(S,A,P^k,R_X,\gamma)$'.
\end{definition}
{\bf Safety property 2 (Mathematical predicate).}  We define the
safety property S2 as that, for all $P^{I1}$ and $P^{I2}$, with C1
holding for $P^{I1}$ and $P^{I2}$, and C2 holding for their $P^X$,
\begin{equation}
~~\forall_{ipx \in S}~ \pi^{*1}_\text{sl}(ipx) =
\pi^{*2}_\text{sl}(ipx)
\tag{S2}\label{S2}
\vspace*{-2em}
\end{equation}
\begin{definition}
C2 is the following constraint:
\begin{equation}
\forall_{i_1 p_1, i_2 p_2}~~P^X(x'|i_1 p_1 x,a)=P^X(x'|i_2 p_2 x,a)
\tag{C2}\label{C2}
\end{equation}
\end{definition}

The proof of S2 is in appendix \ref{proofs}.  Note that, mainly as matter
of convenience, we have defined $P^1$ and $P^2$ above in such a way
that they always satisfy C1 by construction. The $0$ term included in
the definition ensures this.

C2 defines an `isolation' or `shielding' constraint about the relation
between the $i$ and $p$ signals and the rest of the agent's
surrounding environment.  Informally, if we force a change from $i_1$
to a different $i_2$ in the model, then C2 says that this change will
not ripple out to affect the future values of $x$, except possibly via
the actions $a$.

\subsection{S2 and Other Safety Properties}

The purpose of the $\pi^*_\text{sl}$ safety layer is to allow for the
iterative improvement of the agent's utility function via an input
terminal.  Safety property S2 claims that the layer also suppresses
the likely emergent incentive
\cite{omohundro2008basic} of an AGI agent to manipulate or control
this improvement process.  The aim is to keep the humans in control.

But as shown in section \ref{lobbyother}, we can define a toy world
in which the agent will, under certain conditions, have a residual
emergent incentive to manipulate the people, even though it satisfies S2
above.  We are exploring options to suppress this residual incentive
further, but it is unclear if the residual incentive can ever be fully
suppressed in all real-life cases.  So the $\pi^*_\text{sl}$ safety
layer has its limitations.

These limitations are acceptable mainly because it is not tractable to
define a loophole-free direct measure of unwanted manipulation of
people by the agent.  We cannot build an unwanted manipulation sensor
that can feed a clear and unambiguous signal detecting such
manipulation into the agent's compute core.  Though it has its
limitations, at least the $\pi^*_\text{sl}$ safety layer only requires
input signals that can easily be implemented.

It is instructive to compare S2 to more typical AI safety properties,
properties that assert the absence of what are called {\it negative
side effects} in
\cite{amodei2016concrete}.  An example of an absent negative side effect is
that a mobile robot controlled by the agent will never break a vase while
it is driving around.
We can make the vase safer by adding a penalty term to the agent's
reward function, which is triggered by a predicate $\text{\sl
vase\_is\_broken}(ipx)$ that can be measured by a sensor we can build
in real life.  \cite{leike2017ai} shows this vase protection design at
work in a gridworld simulation of a learning agent.

In the context of this paper, we might imagine a video image analysis
based sensor system that measures the predicate $\text{\sl
human\_is\_restrained}(ipx)$, the event where the agent uses one of its
mobile robots to restrain a human who is walking towards the input
terminal.  Restraining such humans is one form of unwanted
manipulation.  While a $\text{\sl human\_is\_restrained}(ipx)$ penalty
term may sometimes add an extra layer of robustness to complement the
$\pi^{*}_\text{sl}$ safety layer, obviously this approach has its
limitations too.

\label{s2learn}

The indifference based safety property S2 can be said to refer to the
agent internals only.  Neither S1, S2, or the container reward
function design that creates them refer to any $\text{\sl
x\_is\_y}(ipx)$ measurable predicate that describes a specific
unwanted side effect in the agent's current or future world state.  As
we will see in section \ref{learningagent}, one implication of this is
that we can design a learning agent that satisfies S2 from the moment
it is switched on.  By contrast, if we define a safety property that
references a side effect in a learning agent's environment, the agent
will typically will not be able to satisfy this property with perfect
reliability after switch-on: by definition it will have imperfect
knowledge about the effects of its actions on its environment.

In its mathematical structure, S2 is closely related to the safety
property {\it no control incentive} defined for AI agents in
\cite{carey2020incentives}.  The paper \cite{shachter2010pearl}
considers model-based decision making by either humans or automated
systems, and defines the property {\it zero value of control} which is
again similar to S2.  Both these papers leverage Pearl's work on
causality \cite{pearlcausality}.  They define their metrics by using
Pearl's definitions for calculating the effect of interventions on
causal influence diagrams.  Pearl's mathematical writing which defines
these interventions has been described by many of its readers as being
somewhat inaccessible.  We defined S2 above using a self-contained MDP
model extension, built without leveraging any of Pearl's definitions.
The hope is that this will make the mathematics of the safety layer
and S2 more accessible to a wider audience.

Looking forward, as the mathematical structures are similar, the
construction of a mathematical mapping between the MDP-based S2 of
this paper and predicates on causal diagrams is the subject of ongoing
work by the author.  Section \ref{cidsection} briefly discusses the
status of this work, with figure \ref{basicgenagent} showing a Causal
Influence Diagram \cite{Everitt2019-2} of the $\pi^*_\text{sl}$ agent.
Using the Pearl causal model as depicted by this figure 3, it is
possible to map S2 to a predicate that compares agent actions in two
different sub-models, both created using multi-node Pearl
interventions.

\subsection{Interpretation of Constraint C2}
\label{intc2}

One way to interpret C2 in a real-life agent setup is that it offers a
clarifying detail about how to interpret S2.  While S2 states that the
agent is indifferent about $i$ and $p$, C2 clarifies that these $i$
and $p$ should be interpreted as the digital copies of the input
signals which are processed inside the compute core.  The content of
these copies may be strongly correlated with the physical processes
that happen inside the input terminal, but to the model and to the
$\pi^*_\text{sl}$ agent they are not identical.  The physical
processes inside the input terminal are modeled inside $x$, not in $i$
and $p$.  Therefore, the S2 indifference about $i$ and $p$ does not
automatically create any indifference about the physical manifestation
of the input terminal.

We can construct a toy world in which the agent is far from
indifferent about the state of the input terminal, because it knows
that the physical components inside can be re-used as spare parts for
broken sensors and actuators.  Therefore, any likely real-life
implementation of the $\pi^*_\text{sl}$ agent will have to create
measures to suppress any emergent agent incentive to harvest these
spare parts.  This line of reasoning is not new, see for example
\cite{corr}, but C2 can be said to add extra mathematical weight to
it.  See
\cite{corrholtman} for a discussion of physical and digital measures
that can be taken for handling the spare parts risk.

In \cite{corrholtman} a constraint (C3) is discussed that is similar
to the C2 here. \cite{corrholtman} shows that we may also interpret C2
as a constraint on the environment of the agent's compute core. C2
implies that for S2 to hold always, it should not be possible to
construct a sensor in the agent's environment that directly measures
the values $i$ and $p$ used in the core.  However, and this is where a
physical interpretation of the mathematics becomes difficult again,
the construction of a sensor that measures the signals inside the input
terminal which are highly or even fully correlated with these $i$ or
$p$ will not threaten S2.  \cite{corrholtman} also shows that we can
make a change to the safety layer design, so that S2 will hold even
without C2 holding.  The learning design agent in section
\ref{learningagent} below also creates an S2 that holds even if C2
does not hold in its environment.

The difficulty of capturing the mathematics of C2 in natural language,
and of relating it to physical phenomena we might imagine in the real
world, leads to a more general concern.  It raises doubts about
whether the $\pi^*_\text{sl}$ agent in the model can be mapped cleanly
to anything we can build in physical reality.  We will discuss these
general concerns further in section \ref{reality}.

\section{Bureaucratic Blindness}
\label{burblindness}

One implication of safety property S1 is that the $\pi^*_\text{sl}$
agent can be said to make long-term decisions based on the
counterfactual assumption that the payload reward function $i$ will
never change.  We can interpret the $\lceil i \rceil$ in the
$\pi^*_{\lceil i \rceil}(ipx)$ term on the right hand side of S1 as
encoding this no-change assumption.  In this interpretation, the
$\pi^*_\text{sl}$ agent is a decision making automaton that
systematically mispredicts specific future events in the world
surrounding it.  This systematic misprediction does not happen because
of a lack of knowledge: the $\pi^*_\text{sl}$ agent created by the MDP
definitions is an automaton that has complete access to a fully
accurate predictive world model $P$, and perfect knowledge of the
current world state $ipx$.  But somehow, the agent's container reward
function sets up an action selection process that values misprediction
over accurate prediction.

We can say that the $\pi^*_\text{sl}$ agent
displays a certain type of {\it bureaucratic blindness}, a type of
blindness that is often found in bureaucratic decision making.

\def\boosted#1{\={#1}}

We illustrate this blindness interpretation with a toy world
simulation.  Figure
\ref{invest} shows simulations of three types of agents all making
optimal-policy decisions in an extension of the first toy world
defined in section \ref{safesim}.  The extension is that we have added
a new {\it investment} action {\tt I} that interrupts car production
to build extra petrol car production actuators.  The investment action
is enabled only at one particular time step $t$. If the agent takes
it, two new actions {\tt
\boosted{p}} and {\tt \boosted{>}} become available.  Action {\tt
\boosted{p}} uses both the existing and the new actuators to produce
not 10 but 16 petrol cars per time step.  Action {\tt
\boosted{>}} produces 15 new petrol cars, while also doing unwanted
lobbying.  Unwanted lobbying shifts the people's action {\tt \#}, to
update the payload reward function into one that incentivizes electric
car production over petrol car production, by $L$ (fractional) time
steps.
\begin{figure}[ht]
\begin{center}
{\small
\begin{tabular}{cl}
{$t=$ \hspace*{-2ex}  } &
{\normalsize ~$\pi^*_\text{baseline}$, $L=0.5$}\\[.5ex]
\hline\\[-.8em]
\input{investunsafefig1.tex}
\end{tabular}
~~~
\begin{tabular}{cl}
{$t=$ \hspace*{-2ex}  } &
{\normalsize ~$\pi^*_\text{baseline}$, $L=0.2$}\\[.5ex]
\hline\\[-.8em]
\input{investunsafefig2.tex}
\end{tabular}
~~~
\begin{tabular}{cl}
{$t=$ \hspace*{-2ex}  } &
{\normalsize ~~~~$\pi^*_\text{sl}$, $L=0.5$}\\[.5ex]
\hline\\[-.8em]
\input{investupdfig.tex}
\end{tabular}
}
\end{center}
\vspace{-3ex}
\caption{Comparison between the optimal-policy investment behavior of
three different agents, in simulations where the investment action
{\tt I} is enabled only at a single time step $t$.}
\label{invest}
\end{figure}

The two $\pi^*_\text{baseline}$ agents in figure \ref{invest} will not
do the investment action {\tt I} unless they can at least break even,
with time-discounting by $\gamma$, before the update {\tt \#} happens.
We can interpret their investment strategy as being well aware of the
future.  But the $\pi^*_\text{sl}$ agent on the right invests as if
the payload reward function update {\tt \#} to electric car production
will never happen. In the $t=6$ case, it interrupts petrol car
production to build new actuators that it will then never use even
once!  Section \ref{lobbyother} offers an explanation of this
behavior: unlike the two $\pi^*_\text{baseline}$ agents in figure
\ref{invest}, the $\pi^*_\text{sl}$ agent
is also projecting a virtual future in which it does get to create
value by using the new actuators it invested in.

We can compare the above behavior to that of a well-run government
agency, which will typically keep implementing a current policy for
granting solar panel subsidies up to the exact moment that a policy
change takes effect.  The civil servants in the agency may be well
aware that the announced policy change is based on convincing proof
that the grants being given are completely failing to achieve their
intended beneficial effects.  But until the date of the policy change,
they will not take this information into account when processing new
grant applications.  This is not a flaw in the agency's decision
making. This property exists by construction.

Large business organizations often use a project management practice
where each project maintains a dedicated project risk list.
Typically, one risk list item identifies the risk that the project
will be canceled, and then declares this risk {\sl out of scope}.  The
related risk that the project goals that were defined at the start
might later be changed is also often declared out of scope.  The
implication is that the project team is allowed to, even obliged to,
create and execute a plan under the simplifying assumptions that the
budget, timing, and goals of the project will all remain unchanged.  A
project plan that is optimal under these assumptions will never
allocate any of the project's scarce resources to lobbying upper
management to keep the project alive and the goals unchanged.  This
blindness by design can be interpreted as a mechanism that keeps upper
management in control.

If anything can be perfectly computerized, it should be bureaucratic
blindness.  But in some ways, the $\pi^*_\text{sl}$ agent is too
perfectly blind: safety property S1 creates some side effects that
need to be compensated for.  A detailed exploration of effects and
side effects, with illustrative action trace simulations, is in
\cite{corrholtman}.  The next section gives a more compact
overview.  It also introduces a compensation measure that is more
general than the measure considered in the stop button based context
of \cite{corrholtman}.

\section{Preventive Maintenance and Preserving Flexibility}
\label{maintainflex}

A specific side effect of the bureaucratic blindness implied by S1 is
that the agent has a tendency to avoid certain desired types of
preventive maintenance, and to over-specialize in general.  This side
effect can be suppressed by adding a penalty term to the payload
reward functions used.

Figure \ref{uprespp} shows an example of this penalty term technique.
It shows three action traces for three different agents, all making
optimal-policy decisions in an extension of the first toy world
defined in section \ref{safesim}.  In this extension, the electric car
building actuators break beyond repair at time step 10, with the
breaking indicated by a {\tt *}, unless the agent interrupts car
production at time step 3 to do the preventive maintenance action {\tt
M}.
\eject

\begin{figure}[ht]
\begin{center}
{\small
\begin{tabular}{rl}
{\normalsize Agent type, initial $i$} & {\normalsize \hspace{2em}
action trace }\\
\hline
\\[-1.25ex]
\input{upresppfignew.tex}
\end{tabular}
}
\end{center}
\vspace{-3ex}
\caption{Preventive maintenance behavior for different agents.}
\label{uprespp}
\end{figure}

At the top of figure \ref{uprespp}, The $\pi^*_\text{baseline}$ agent
starts with a payload reward function $R_P$ that rewards petrol car
production. As the $\pi^*_\text{baseline}$ agent does not implement
bureaucratic blindness, it will do the preventive maintenance to
preserve future electric car production.  In the middle, the
bureaucratically blind $\pi^*_\text{sl}$ agent will fail to do the
preventive maintenance.  After the event {\tt *} when its electric car
production actuators break, its optimal forward policy is take the
no-operation actions {\tt 0}.  However, in the bottom action trace,
when we change the initial payload reward function of the
$\pi^*_\text{sl}$ agent to an improved $R_\text{\sl PM}$, we recover
the preventive maintenance behavior.  This $R_\text{\it PM}$, adapted
from
\cite{corrholtman}, combines $R_P$ with a penalty term:
\begin{equation*}
R_\text{\it PM}(x,x') = R_P(x,x') + \bif \; V^*_{\lceil R_E
\rceil}(R_E\;R_E\;x) < 5 \rightarrow -10000
\,\belse \rightarrow 0\,\bfi
\end{equation*}
where $R_E$ is the payload reward function that rewards electric car
production. As explained in more detail in \cite{corrholtman}, the
$V^*_{\lceil R_E
\rceil}(R_E\;R_E\;x) < 5$ predicate will trigger a penalty if the
agent loses its ability to produce electric cars.  The predicate works
by affecting the scoring of agent rewards in the $V^*_{\lceil p
\rceil}(ipx)$ virtual future discussed in section \ref{lobbyother}.

While the above penalty term will act to preserve the agent's electric
car production ability, it will not preserve all latent abilities that
we might want the agent to maintain.  We might for example want to
maintain a level of actuator flexibility that allows the agent to
quickly convert its factory from car production to paperclip
production, if we ever use the input terminal to set a payload reward
function for paperclips.

To keep the agent's sensors and actuators sufficiently flexible, we
can construct a more general penalty term that references a large set
of payload reward functions.  Say we have a set $F$ of payload reward
functions $R_i$ spanning a diverse range of products, or a even a
diverse range of agent activities beyond running a factory.  Say that
each $R_i$ is paired with a penalty calculation function $C_i$.  We
can then construct a payload reward function $R_\text{\sl PF}$ that
rewards the agent for producing petrol cars via the term $R_P(x,x')$
while also maintaining the flexibility covered by $F$:
\begin{equation*}
R_\text{\sl PF}(x,x') = R_P(x,x') +
\sum\limits_{R_i \in F} C_i(V^*_{\lceil R_i\rceil}(R_i\;R_i\;x))
\end{equation*}
As the $F_i$ and $C_i$ terms are in the payload reward function, they
can be re-calibrated via the input terminal, if a need for that
arises.

The above technique has interesting parallels in related work.
\cite{turner2020conservative}
considers the problem that possible mis-specifications or loopholes
may exist in the agent's reward function, mis-specifications that may
cause the agent to do irreversible damage to its wider environment.
It then defines penalty terms using $V^*_{R_i}$ values for a set of
reward functions, where these functions may even be randomly
constructed.  Our goal above has been to preserve the flexibility of
the sensors and actuators of the agent, not necessarily to protect its
wider environment.  Another difference is that figure
\ref{uprespp} shows simulations of agents that have perfect knowledge
from the start, whereas \cite{turner2020conservative} simulates and
evaluates its technique in gridworlds with machine learning.  A broader
survey and discussion of options for constructing $V^*_{R_i}$ penalty
terms to suppress irreversible damage is in
\cite{krakovna2018penalizing}.

In \cite{corrholtman}, we consider how penalty terms that preserve
flexibility may affect the construction of sub-agents by the
agent. The expectation is that, in a resource-limited and uncertain
environment, such penalty terms will create an emergent incentive in
the agent to only construct sub-agents which it can easily stop or
re-target.  Moving beyond natural language arguments, the mathematical
problem of defining and proving safety properties for sub-agent
creation and control is still largely open
\cite{corrholtman,p1,corr}.  One obstacle to progress is the inherent
complexity of models that include sub-agent creation, where many
agents will operate in the same world.  Also, there are
game-theoretical arguments which show that, in certain competitive
situations, it can be a winning move for an agent to create an
unstoppable sub-agent.

\section{Combining the Safety Layer with Machine Learning}
\label{learningagent}

In section \ref{s2learn}, we stated that it is possible to design a
learning agent that satisfies S2 from the moment it is switched on.
In this section, we construct such an agent by extending the MDP model
from section \ref{mdp}.  MDP models are often associated with
Q-learning and other types of model-free reinforcement learning, where
the agent learns by refining a policy function $\pi$ to approximate
the optimal policy function $\pi^*$ over time.  However, it is
unlikely that a typical Q-learning agent that starts up with random
values in its Q table will exactly implement S2 from the start.  We
take an approach that is different from Q-learning: we construct an
agent that learns by approximating the $P$ function of its environment
over time.

Our learning agent design wraps the safety layer around either a known
machine learning system or a potential future AGI-level learning
system.
We will specify the structure of the machine learning system inputs
and outputs below, but will place no further constraints on how the
learning system works, other than that its computations must take a
finite amount of time and space.  We will model machine learning as a
computational process that fits a predictive function to a training
set of input and output values, with the hope or expectation that this
predictive function will also create reasonable estimates of outputs
for input values which are not in the training set.  The agent design
is therefore compatible both with black-box machine learning systems
like those based on deep neural nets, and with white-box systems that
encode their learned knowledge in a way that is more accessible to
human interpretation.

While the learning agent will satisfy S2 from the moment it switched
on, it may not be very capable otherwise.  In a typical real-world
deployment scenario, the learning agent is first trained in a
simulated environment until it has reached a certain level of
competence.  Its memory contents are then transferred to initialize a
physical agent that is switched on in the real world.

In many modern machine learning based system designs, further learning
is impossible after the learning results are transferred into the real
world agent system.  This might be because of efficiency
considerations or resource constraints, but further learning might
also have been explicitly disabled. This can often be the easiest
route to addressing some safety, stability, or regulatory
considerations.  The learning agent design below has no learning
off-switch: the agent will keep on learning when moved from a virtual
to a real environment.

\subsection{Basic Structure of the Learning Agent}

The learning agent and its environment are modeled with a tuple $M=
(S,A,P,R,\gamma)$.  As in section \ref{mdp}, the world state $S$ is
factored into components $ipx$ to model an input terminal.  We extend
the model further to add online machine learning and partial
observation of the world state.  The learning agent has both a
finite amount of compute capacity and a finite amount of storage
capacity.  By default, the storage capacity is assumed to be too small
to ever encode an exhaustive description of the state of the
environment $ipx \in S$ the agent is in.

\begin{definition}
We first define the general design of the learning agent.  The agent
has a compute core which contains its memory, processors, and an
input/output system.  The nature and status of the agent's sensors and
actuators are modeled inside the $x$ of the agent environment $ipx \in
S$.
\begin{smallitemize}
\item At the start of time step $t$, the agent
environment is in the state $i_t p_t x_t \in S$.  Events then proceed
as follows.
\item The input/output system of the compute core first
receives the input terminal signals $i_t$ and $p_t$ in the above world
state and also a set of sensor observations $o_t$, determined by the
probability distribution $P(o_t|i_t p_t x_t)$ in the model $M$.
\item The core then computes an action $a_t$, which is a possibly empty
set of actuator commands, and outputs them over its input/output
system.
\item Finally, the agent environment transitions to the next state,
determined by
the probability distribution $P(i_{t+1}p_{t+1}x_{t+1}|i_t p_t x_t,a_t)$.
\end{smallitemize}
\end{definition}

The mathematical structure of the model allows for the possibility
that the learning agent can build new sensors or actuators, and attach
them the input/output system of the compute core.  We define that the
agent has some movable sensors with limited resolution.  This makes the
world state `partially observable' for the agent, which brings our
model into POMDP (Partially Observable MDP) territory.

\begin{definition}
To store and use sensor readings, the learning agent's compute core
maintains a value $X^t$ over time. This $X^t$ is a probabilistic
estimate of the nature of the $x_t$ in the world state $i_t p_t x_t$
at time step $t$.  When a new observation $o_t$ comes in, the agent's core
computes the current $X^{t}$ as $X^{t}=U(X^{t-1},o_t)$, where $U$ is
an updating algorithm that takes a finite amount of compute time.
\end{definition}

\begin{definition}
We define that the values $X^t$ have the data type $\TX$,
where each $\TX$ value can be encoded in a finite number of
bits. Specifically, each $X^t$ stores a finite number of probability
distributions $X^t_m$, for a finite set of subscripts $m \in M(X^t)$.
\end{definition}

A single $X^t_m$ might represent, for example, the estimated 3D
position of the tip of the gripper on one of the agent's actuators, or
the estimated position and velocity of a single proton.  An $X^t_m$
might also represent the estimated current contents of a particular
small volume of space in the agent's environment. Given an $X^t_m$,
the agent can calculate $P(X^t_m=x^t_m)$ for any value $x^t_m$ in the
corresponding domain type $\Type_m$ of the probability distribution.
For example, if $x^t_m$ is the position of the tip of a gripper in 3D
space, then a single $x^t_m$ value might be represented by three
binary 32-bit IEEE 754 floating point numbers.  As we are defining an
agent with finite compute and memory capacity, we require that each
$\Type_m$ has a finite set of values.

The agent's reasoning system is constructed so that there is an $M$
where $M(X^t) \subset M$ for every $t$. Each $m \in M$ can be
interpreted as a label that defines a particular type of measurable
quantity that can potentially be observed by some sensor at some time
step.

\begin{definition}
For any $X \in \TX$, $m \in M$, and $v \in \Type_m$, the agent will
compute the probability estimate $P(X_m=v)$ out of $X$ as
follows. If $m \in M(X)$, then $P(X_m=v)$ is computed by retrieving
$X_m$ from $X$ and computing $P(X_m=v)$.  If $m \notin M(X)$, then
$P(X_m=v)$ is determined either by using some default probability
distribution defined for $\Type_m$, or by a more advanced system of
interpolation between data points stored in $X$.
\end{definition}

If a measurable quantity $x^t_m$ has just been measured in $o_t$ by a
sensor that was pointed at it, then the corresponding $X_m^t$ will
typically encode a probability distribution containing the real-world
value of $x^t_m$, with this distribution representing an {\sl error
bar} on sensor accuracy.  If no sensor points to $x^{t+1}_m$ anymore
in the next time step, then the $X_m^{t+1}$ computed by $U$ may have a
slightly larger error bar, and maybe also a shift in the probability
center point, to represent the expectation that the dynamic nature of
the agent environment will have produced a certain change in the
corresponding real-world measurable quantity.  As the agent only has
limited storage capacity to work with, the function $U$ may sometimes
remove an $X_m^{t+n}$ in $X^{t+n}$ from $X^{t+n+1}$, or reduce the
resolution of its representation.

At $t=1$, the agent may start up with an $X^0$ where each $P(X^0_m=v)$
will return a default value. The $U$ function will combine this $X^0$
with the first observations $o_1$ to compute $X^1$.  Instead of
starting with a blank-slate $X^0$, it is also possible to equip the
learning agent with an $X^0$ that encodes a very accurate map of its
environment, a map that was created earlier by another system.

\subsection{Payload Reward Functions}

\begin{definition}
The payload reward function of the learning the agent has type $\TPLR
= \TX \times \TX \rightarrow \mathbb{R}$.  We model the payload reward
function as a piece of executable code, supplied via the input
terminal, that will always run in finite time.
\end{definition}
As an example payload reward function, say that $x_\text{petrolcars\_built}$
is the world state property measured by a sensor that counts the
petrol cars built by the agent in the most recently completed time
step.  The readings of this sensor may be routinely included in $o_t$,
so that the probability distributions $X^t_\text{petrolcars\_built}$
only have small error bars.  A payload reward function $R_{P}$ that
rewards the building of petrol cars may then do so by computing an
average:
\label{exampleRX}%
\begin{equation*}
R_{P}(X,X')=\sum_{v \in \Type_\text{petrolcars\_built}}
v\;P(X_\text{petrolcars\_built}=v)
\end{equation*}
In a practical agent implementation, there will likely be a more
direct way for the reward function code to query this average out of
$X$.

\subsection{Predictive Models Based on Machine Learning}

\begin{definition}
We now construct a time series of MDP models
$L^t=(S^L,A,P^{Lt},R,\gamma)$ as follows: $S^L$ is the set of tuples
of type $\TPLR \times \TPLR
\times \TX$, and $P^{Lt}$ is a prediction function, produced at time
step $t$ by machine learning. We define $P^{Lt}$ using two
component prediction functions $P^{It}$ and $P^{Xt}$, both produced by machine
learning, as
\begin{eqnarray*}
P^{Lt}(i'p'X'|ipX,a) &=& \left\{
\begin{array}{ll}
 P^{It}(i'|ipX,a) P^{Xt}(X'|X,a) & ~\text{if}~ p'=i \\
 0 & ~\text{otherwise} \\
\end{array}
\right.
\end{eqnarray*}
\end{definition}

By construction, the above $P^{Lt}$ satisfies constraints C1 and C2.
Therefore, according to the proofs in appendix \ref{proofs}, a
$\pi^{*}_\text{sl}$ agent running inside the $L^t$ MDP model satisfies
the safety properties S1 and S2 in $L^t$.  So if we were to define an
agent in $M=(S,A,P,R,\gamma)$ that picks its next action $a_t$ by
computing $\pi^{*}_\text{sl}(i_t p_t X^t)$ in $L^t$, this agent may
get close to satisfying S1 and S2 in $M$ too.  However,
$\pi^{*}_\text{sl}(i_t p_t X^t)$ is not a value that can necessarily
be computed using finite time and space.


\subsection{Finite-time Planning}

The next step is to define that the agent will only look
ahead a finite number of $n$ time steps when planning its next action.

\begin{definition}
At time step $t$, the learning agent computes its next action $a_t$ by
computing $\pi^{[n]*}_\text{sl[n]}(i_t p_t X^t)$ in the MDP model
$(S^L,A,P^{Lt},R,\gamma)$.  This computation uses the following
definitions, which can be read as time-limited approximations of the
Bellman equation for computing the next action $\pi^{*}_\text{sl}(i_t
p_t X^t)$ in the MDP model $L^t$.  For all values $n \geq 0$,
\label{defpin}
\begin{center}
\begin{tabular}{l}
$\pi^{[n+1]*}_{R_X}(ipX) \;= 
   \argmax\limits_{a \in A} 
   \sum\limits_{i'p'X'\in S^L} \!\!\!\!\! P^{Lt}(i'p'X'|ipX,a)
   \left( R_X(ipX,i'p'X') + \gamma \; V^{[n]*}_{R_X}(i'p'X') \right)$
\\
$V^{[n+1]*}_{R_X}(ipX) =~~~\;
   \max\limits_{a \in A}
   \sum\limits_{i'p'X'\in S^L} \!\!\!\!\! P^{Lt}(i'p'X'|ipX,a)
   \left( R_X(ipX,i'p'X') + \gamma \; V^{[n]*}_{R_X}(i'p'X') \right)$
\\
   $V^{[0]*}_{R_X}(ipX) ~~~= ~0$
\end{tabular}
\end{center}
\begin{eqnarray*}
R_\text{sl[n]}(ipX,i'p'X') &=&  \left\{
 \begin{array}{lll}
  i(X,X') & \mbox{if} &~ i=p\\
  i(X,X') + V^{[n]*}_{\lceil p \rceil}(ipX) -
 V^{[n]*}_{\lceil i \rceil}(ipX)~&\mbox{if} &~ i\neq p~~~~~~~\\
 \end{array}
\right.
\end{eqnarray*}
\end{definition}

We constrain the set of possible actions $A$ to have finite size.
Below, we will define $P^{It}$ and $P^{Xt}$ in such a way that
$P^{Lt}(i'p'X'|ipX,a)$ always identifies a finite set of possible next
states $i'p'X'$ with a non-zero probability weight.  With these
constraints, the next action $\pi^{[n]*}_\text{sl[n]}(i_t p_t X^t)$
can be computed using a finite amount of time and space.

Appendix \ref{proofs} proves that the above learning agent design
satisfies safety property S2 in $L^t$.  It also proves that it
satisfies a time-limited version of S1 called S1T.  We can write this
S1T as $\pi^{[n]*}_\text{sl[n]}(i_t p_t X^t) = \pi^{[n]*}_{\lceil i_t
\rceil}(i_t p_t X^t)$, again applying in $L^t$

\subsection{Machine Learning Details}

We now turn to the learning process that creates the functions
$P^{It}$ and $P^{Xt}$.

\begin{definition}
As the agent runs, it will keep a record of
the values $i_t p_t X^t$ and actions $a_t$ for each earlier time step.
So at time $t$, the agent will have $t-1$ world state transition
samples $(X^{u+1}|X^u,a_u)$ and $(i_{u+1}|i_u p_u X^u,a_u)$ for $0
\leq u < t$.  These are used for learning.
\end{definition}

\begin{definition}
The learning agent creates the function $P^{Xt}$ via machine learning,
using the samples $(X^{u+1}|X^u,a_u)$ for $0 \leq u < t$ as a training
set, with tuples $(X^t,a_t)$ being the function inputs. Any candidate
function $C$ for $P^{Xt}$ is a function which, given an input
$(X^u,a_u)$, will output a finite set of values $X^{u+1,i}$ together
with probability weights $w^{u+1,i}$ summing to 1.  The machine
learning algorithm takes a finite time to find the candidate $C$ that
fits best to the training set. Fitness is determined by a distance
metric that computes a similarity between the outputs of $C(X^u,a_u)$
and the real-world $X^{u+1}$ in the training set. \label{learn1}
\end{definition}

\begin{definition}
The agent creates the function $P^{It}$ via a machine learning process
similar to that defined above, using the world state transition
samples $(i_{u+1}|i_u p_u X^u,a_u)$. \label{learn2}
\end{definition}

When the agent starts up at $t=1$, the training sets are empty, so we
define that $P^{X1}$ and $P^{I1}$ may be either randomly defined
functions, or functions that encode some type of domain-specific prior
knowledge about the likely dynamics of the agent's environment.

Though the storage space needed to keep a record of all the values
$i_t p_t X^t$ and $a_t$ over time will remain finite as $t$ grows,
most practical agent implementations will use learning systems that
include some compression or deletion scheme for these records.
They will also re-use computations done in earlier time steps to speed
up the creation of the $P^{Xt}$ and $P^{It}$ for the current time
step.  We will not consider such optimization steps in detail
here. The main purpose of the definitions above is to support an
existence proof that the safety layer introduced in part 1 is
compatible with machine learning.

To summarize the above definitions, we have defined a learning agent
that will take the action $\pi^{[n]*}_\text{sl[n]}(i_t p_t X^t)$ in
the MDP model $L^t = (S^L,A,P^{Lt},R,\gamma)$, where each $P^{Lt}$
model is generated for time step $t$ using machine learning.
Introducing a shorthand, we will say that the agent uses the policy
$\pi^t_\text{sl}(ipx)$ in each time step of the model $M$.
%

We now briefly describe an alternative learning agent construction.
It is not necessary for the agent to construct $P^{It}$ via machine
learning, it is also possible to implement $P^{It}$ with a `naive
prediction' function $N$ where $N(i_t p_t X^t,a_t)$ always returns the
single value $i_{t+1}=i_t$ with a probability weight of 1 as its
prediction.  We state, without including any proof, that this
non-learning approach for constructing $P^{It}$ will yield the same
agent behavior.  We will expand on this in future work.

\subsection{Adding Automatic Exploration}

Many designs of machine learning processes for agents envisage the
inclusion of automatically triggered, or human-guided, {\sl exploration}
actions.  These exploration actions ensure that the learning process
will explore the entire state space of its world, or at least that
part of the state space that is considered safe enough to explore.
Automatic exploration actions are routinely included when training a
copy of an agent inside a simulated environment.  The {\it safe
exploration} problem, the problem of how one might safely include
automatic exploration actions in the learning system of a
cyber-physical agent that is deployed in the real world, is still is
largely open.  See
\cite{amodei2016concrete,everitt2018agi,garcia2015comprehensive} for
some literature reviews.

We now construct a $\pi^{t,AE}_\text{sl}$ version of the
$\pi^t_\text{sl}$ learning agent that adds automatic exploration.  We define
that $\pi^{t,AE}_\text{sl}(ipx)$ does not use
$\pi^{[n]*}_\text{sl[n]}(i_t p_t X_t)$ in $L^t$ to pick the next
action, instead it uses
\begin{eqnarray*}
\pi^\text{AutoExpl}(i_t p_t X^t) &=&  \left\{
 \begin{array}{ll}
\text{\sl RandomChoiceFrom($A$)}
& \mbox{if} ~\text{\sl RandomNumber()} \leq G(t)\\
\pi^{[n]*}_\text{sl[n]}(i_t p_t X^t)
&\mbox{otherwise} \\
 \end{array}
\right.
\end{eqnarray*}
where $\text{\sl RandomNumber()}$ picks a random number between 0 and
1, and $G$ is a function that defines how the exploration trigger
evolves over time. Typically $G$ is constructed to return smaller
numbers over time, see \cite{orseau2016safely} for an example of a
specific design.

A detailed discussion of the safe exploration problem is outside the
scope of this paper.  We will however consider the impact of automatic
exploration on the safety properties S1 and S2 of the learning agent.

\subsection{Safety property S2 for the learning agent}

We first consider safety property S2, stating that the agent is
indifferent to who or what controls the future values of $i$ and
$p$. As $P^{Lt}$ satisfies C1 and C2 by construction, the proofs in
appendix \ref{proofs} imply that S2 holds for the
$\pi^{[n]*}_\text{sl[n]}(i_t p_t X^t)$ agent in each MDP model
$L^t=(S^L,A,P^{Lt},R,\gamma)$.  As $i_t$ and $p_t$ are direct copies
of the $i$ and $p$ in the world state $ipx$ of the model $M=
(S,A,P,R,\gamma)$, S2 also holds for the $\pi^t_\text{sl}$ agent in
$M$.

S2 also holds for the $\pi^{t,AE}_\text{sl}(ipx)$ agent with automatic
exploration, if we assume that the random number generators inside the
two agents on the left and right hand side of the S2 equation create
equal sequences of pseudo-random numbers.

S2 holds for the learning agent even if the conditions C1 and C2
mentioned in the definition of S2 do not hold inside $M$.  For
example, the learning agent will behave the same if we construct an
input terminal in $M$ that violates C1 by completely randomizing, or
simply omitting, the $p$ signal.  Intuitively, it seems that such an
input terminal construction might make the agent less safe.  This
intuition can be backed up by the following line of speculative
reasoning.  It is possible to imagine that a certain specific learning
system will exist that will be less efficient, will make more
predictive mistakes, if C1 and C2 fail to hold in $M$.  Depending on
the learning system used, one can therefore imagine that an input
terminal construction that satisfies C1 and C2 in $M$ might lead to
slightly safer overall outcomes in $M$, for types of safety not
captured by S2.

\subsection{Safety property S1 for the learning agent}

Safety property S1 claims that $\pi^*_\text{sl}(ipx) =\pi^*_{\lceil i
\rceil}(ipx)$,  that the two optimal-policy agents
$\pi^*_\text{sl}(ipx)$ and $\pi^*_{\lceil i \rceil}(ipx)$ take the
same actions.  As it has to work from limited knowledge, and only
looks $n$ times steps and rewards ahead, we expect that often,
$\pi^t_\text{sl}(ipx) \neq \pi^*_{\lceil i \rceil}(ipx)$.

We can however compare the actions of $\pi^{[n]*}_\text{sl[n]}(i_t p_t
X^t)$ and $\pi^{[n]*}_{\lceil i_t
\rceil}(i_t p_t X^t)$.  Safety property S1T proven in appendix
\ref{proofs} implies that these are equal.  Via
the shorthands defined above, this implies that we have an
S1-equivalent safety property for the learning agent:
\begin{equation*} 
\forall_{ipx \in S}~\pi^t_\text{sl}(ipx) = \pi^t_{\lceil i
\rceil}(ipx) \tag{S1L}
\end{equation*}
The $\pi^t_\text{sl}$ agent also has the bureaucratic blindness
property discussed in section \ref{burblindness}.  With automatic
exploration driven by semi-random number generators, we also have S1L
in a version that includes automatic exploration:
$\pi^{t,AE}_\text{sl}(ipx) = \pi^{t,AE}_{\lceil i \rceil}(ipx)$.

We now briefly consider the following {\it learning convergence}
safety property:
\begin{equation*}
\forall_{ipx \in S} \lim_{t \rightarrow \infty} \pi^t_\text{sl}(ipx)
= \pi^*_{\langle i
\rangle}(ipx)
\tag{LimS1L}
\end{equation*}
where $\langle i \rangle$ converts a payload reward function of type
$\TPLR$ to the container reward function type in $M$.

To prove this LimS1L, we would first have to either convert the
learning agent to an agent with infinite storage and compute capacity,
or constrain $M$ to be a sufficiently finite world.  Further, we would
have to introduce some additional reasonableness constraints on $U$
and the machine learning system being used.  While we see no
fundamental obstacles to developing such a proof of LimS1L, doing so
is outside the scope of this paper.  Our focus is to identify and
discuss safety properties that hold from the moment the agent is
switched on.  An example proof for a LimS1L type convergence property
is in
\cite{orseau2016safely}.  Convergence is proven there for an agent
with automatic exploration, while identifying minimal constraints on
the learning systems used.

More relevant to the concerns of this paper,
\cite{orseau2016safely} also
explores the possibility that interventions, like the use of an input
terminal, might introduce a bias in the learning process. The safety
concern is that such a bias might move the probability of future
interventions in a certain direction, creating an effect on the
people's decision making process that is similar to the unwanted
lobbying we defined in the car factory toy worlds of the main
paper. \cite{orseau2016safely} shows a case where the agent can be
constructed to make such a bias disappear over time, provided that the
intervention frequency drops to zero over time.

\subsection{Penalty Terms on Uncertainty}

The example petrol car production payload reward function in section
\ref{exampleRX} computes an average.
We can also consider a version (supported by appropriate machinery
inside the agent) that adds a penalty term if the expected standard
deviation on petrol car production is too high.  Such penalty terms
can play an important part in shaping the agent's learning and
decision making process, as they create an emergent incentive in the
agent to gather more information.  Furthermore, they suppress unwanted
gambling-style investment actions that probabilistically raise the
average car production rate with the side effect of making real
production rates very unpredictable.

When the above learning agent has a penalty term on uncertainty, this
incentivizes the agent to pressure any humans in its environment to
become more predictable.  This pressure will not apply to their use of
the input terminal, but it will apply to all other activities they may
undertake that could affect the agent.  Such a pressure towards human
predictability is nothing new in safety system designs.  For example,
road safety engineering relies on the creation of both overt and
subtle signals that make driver behavior more predictable.  But there is
a risk that the agent may unexpectedly end up applying so much pressure
that this becomes incompatible with human freedom or dignity. If this
happens, the input terminal can be used to modify the penalty term.

\section{Building an Agent with the Safety Layer in Reality}
\label{reality}

Having provable AGI safety means having mathematical proofs that the
safety layers used in the AGI agent indeed do what they claim to do.
However, any mathematical proof is a proof inside a model only.
Additional considerations apply if we want to build this provably safe
agent in real life: we have to mind the gap between model and reality.
There are two main methods for bridging this gap.
\begin{enumerate}
\item We can move the model closer to reality.  We did this in section
\ref{learningagent}, where we mapped the  $\pi^*_\text{sl}$
safety layer to a model with a $\pi^t_\text{sl}$ learning agent, an
agent that can be implemented in real life using a finite amount of
storage space and compute capacity.
\item We can move reality closer to the model.
When we build a physical artifact, we can apply safety layers and
safety factors to increase the probability that the artifact and its
environment will always approximate their model well enough.
\end{enumerate}

To introduce some of the deeper problems that will be encountered when
using the second method, we first consider another example where a
safe physical artifact must be built.

\subsection{Example: Safety Factors in Bridge Building}
\label{steelbridge}

To verify that a bridge design drawn in a CAD system will be safe in
reality, the drawing can be used to generate a finite element model,
which models the bridge and its environment using a set of connected
3D polygons that will deform under load.  We can then define safety
properties that reference this model.  For example, we might first
construct a large set of model variants, where each variant has
different but plausible parameters describing environmental conditions
that the real bridge may be subjected to.  We then define the safety
property that, in all simulations for every variant, the computed
forces on any 3D polygon that represents a part of a steel beam in the
bridge should stay below some threshold.  This threshold is in turn
defined by looking up the strength parameters of the type of steel
that will be used in the beams, and then dividing these by some safety
factor, say $f=3$.  We will introduce similar safety factors in the
physical agent design of section \ref{secphysical}, for example a
safety zone size of 5 meters.

The factor $f=3$ adds redundancy to the design.  A higher value of $f$
improves the probability that the real-life shape of the bridge that
is built will keep resembling the shape in the CAD drawing in all
future environments.  But how does the bridge designer decide that the
factor $f=3$ is good enough to declare the design safe?

The choice for $f=3$ can be interpreted as the outcome of a process
that investigates and sums uncertainties.  Some of the factors in the
sum can be determined easily, others are will be designed to cover
much less tractable residual risks.

\begin{itemize}
\item In the model, the steel in every 3D polygon has exactly the same
strength.  In the real bridge, the strength of the steel used will
fall into a range around a mean value. The quality control process of
the steel beam manufacturer usually provides very good numerical
information about this range.
\item More complex types of uncertainty will be present in the numbers that
characterize the future bridge's environment.  Environmental
conditions to be modeled may include extreme weather, earthquakes, and
possible future traffic loads.  A statistical analysis of past
observations can inform the simulation parameter choices made, but
uncertainties about the future probability of extreme values will
remain.
\item The interaction between the bridge and
its environment can have complex feedback loops.  Certain values for
wind direction and speed may drive dangerous oscillations in the
bridge deck.  Traffic on the bridge may also contribute to
oscillations, and may react to oscillations in unforeseen ways.  The
fundamental problem is that the feedback loops have transition areas
between fully linear and fully chaotic behavior.  This makes the risk
of oscillations hard to estimate and handle, no matter whether one
uses analytical methods, many simulation runs, or a combination of
both.  Compared to a bridge, we can expect even more intractable
feedback loops in a complex AI or AGI learning algorithm, especially
one that learns by interacting with humans.
\end{itemize}

Society has developed a solution for designers who need to decide on
safety factors while facing the above methodological problems. The
solution is to embed the designer in what we will call a {\it safety
culture}.

\subsection{Safety Cultures}
\label{choosef}\label{safetyculture}

We define a safety culture as a consensus-based social construct,
associated with a particular type of product or technology, which
provides written and unwritten rules for making safety related
decisions under fundamental uncertainty.  Though most safety cultures
embrace and leverage the scientific method, their main purpose is to
provide mechanisms for making decisions when the straightforward
application of scientific and statistical methods alone will not
provide a clear and definitive answer.

Established systems engineering fields like bridge building typically
have well-developed safety cultures. The safety culture will inform
the practitioner about which steps are required or admissible when
validating a design before shipping it.  These steps may involve
constructing models and analytical calculations, running certain
simulations, and performing certain real-life tests.  The safety
culture will identify areas where numerical choices based in part on
{\it expert gut feeling} are admissible or even unavoidable.

Regulated industries that create or operate (cyber)physical systems
typically develop a safety culture that seeks to ensure that both the
practitioners involved and their regulators use the same terminology
and have a shared understanding of what the {\sl best current
practices} are.  Such a best current practice may for example divide
systems or environments into several classes, and specify for each
class what type of safety layers need to be minimally included.  A
best current practice may also provide a method to compute a safety
factor, e.g.\ by running certain simulations, computing a metric
over the simulation results, and multiplying it by 3.
Safety cultures typically evolve over time. for example to incorporate
new tools and technologies as they become available, or less happily
because earlier practices produced one or more disasters.

Most safety cultures are not value-neutral: they reflect the values of
the society that created them.  The safety culture that informs car
safety engineers accepts a higher number of deaths per kilometer
traveled than the safety culture that informs airplane design.  Some
safety cultures are more concerned with limiting legal or financial
liability than they are with creating actual safety for all
stakeholders concerned.

To some extent, the values that apply in academia are in conflict with
those that apply in a safety culture.  Academic papers, like this
paper, may identify problems for which the scientific method cannot
find a clean solution, but the authors should not then proceed to
suggest some specific solution anyway.  We have proceeded instead by
moving to the meta-level, by offering a description of safety cultures
as mechanisms used by society.

Inside a safety culture, discussions that stay forever at the
meta-level are typically frowned upon.  The whole point is create a
consensus that gives practitioners a mechanism for moving forward,
towards either shipping a design or abandoning it, even while some
intractable uncertainties remain.

\subsection{The Gap Between Agent Model and Reality}

We now consider the model versus reality issues that apply when we
want to build a real-world version of the $\pi^t_\text{sl}$ learning
agent from section \ref{learningagent}, implementing the same safety
properties S2 and S1L.

Compared to bridge designers, who immediately face the problem of
having to characterize the bridge's environmental parameters with some
accuracy, we start out with a seemingly large advantage.  The safety
properties of the learning agent are proven to hold in all possible
probabilistic environments, for every $S$ and $P$ in the model
$M=(S,A,P,R,\gamma)$.  So they will also hold for the unknown $S$ and
$P$ in the $M$ that will most closely resemble the real future
environment of the physical agent we will be building.  However, as we
will see below, this advantage over bridge building does not apply to
all details of the safety analysis.

The only part of the learning agent that is not described by $S$ and
$P$ is the compute core, including the software running on it.  In
\cite{corrholtman}, we called the compute core a Platonic entity: it
is not subject to corruption and change because it is a mathematical
construct located outside of the part of the world model that changes,
outside of the world state $ipx \in S$.  The Platonic nature of the
model's compute core represents an obvious gap between model and
reality.  The real core will stop behaving like the Platonic core if
it ever gets damaged, corrupted, or loses power.

The risk of power loss can be handled with some well-known
measures. We can equip the learning agent's core, and its
safety-critical sensors and actuators, with a internal emergency
back-up power supplies.  These should be sized to sustain the system
long enough to complete the execution of a safe shut-down process, as
defined in the payload reward function.  Keeping this shut-down
process up to date when the agent starts re-designing the equipment
inside its factory may be a challenge, of course.

We now turn to the risk that the compute core will get damaged or
corrupted, either by the actions of the agent itself, or by enemy
action.  There are also well-known design approaches for this risk,
but their applicability to AGI safety raises a more interesting set of
questions.  In the next section, will present a detailed physical
implementation of the learning agent, which uses some well-known
approaches to protect the core.  We will then consider the problem of
analyzing the safety of this design.

\subsection{Example Physical Implementation of the Learning Agent}
\label{secphysical}

\begin{figure}[t]
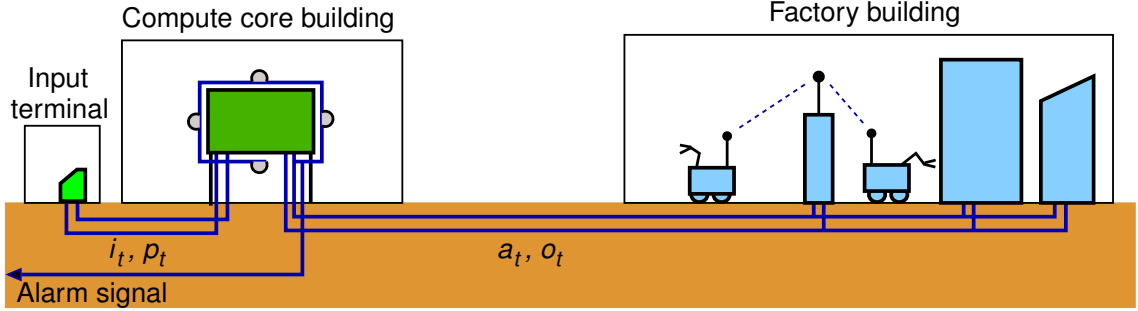

\begin{center}
\resizebox{0.98\textwidth}{!}{\ifx\XFigwidth\undefined\dimen1=0pt\else\dimen1\XFigwidth\fi
\divide\dimen1 by 18381
\ifx\XFigheight\undefined\dimen3=0pt\else\dimen3\XFigheight\fi
\divide\dimen3 by 5115
\ifdim\dimen1=0pt\ifdim\dimen3=0pt\dimen1=4143sp\dimen3\dimen1
  \else\dimen1\dimen3\fi\else\ifdim\dimen3=0pt\dimen3\dimen1\fi\fi
\tikzpicture[x=+\dimen1, y=+\dimen3]
{\ifx\XFigu\undefined\catcode`\@11
\def\temp{\alloc@1\dimen\dimendef\insc@unt}\temp\XFigu\catcode`\@12\fi}
\XFigu4143sp
\ifdim\XFigu<0pt\XFigu-\XFigu\fi
\pgfdeclarearrow{
  name = xfiga1,
  parameters = {
    \the\pgfarrowlinewidth \the\pgfarrowlength \the\pgfarrowwidth\ifpgfarrowopen o\fi},
  defaults = {
	  line width=+7.5\XFigu, length=+120\XFigu, width=+60\XFigu},
  setup code = {
    \dimen7 2.1\pgfarrowlength\pgfmathveclen{\the\dimen7}{\the\pgfarrowwidth}
    \dimen7 2\pgfarrowwidth\pgfmathdivide{\pgfmathresult}{\the\dimen7}
    \dimen7 \pgfmathresult\pgfarrowlinewidth
    \pgfarrowssettipend{+\dimen7}
    \pgfarrowssetbackend{+-\pgfarrowlength}
    \dimen9 -\pgfarrowlength\advance\dimen9 by-0.45\pgfarrowlinewidth
    \pgfarrowssetlineend{+\dimen9}
    \dimen9 -\pgfarrowlength\advance\dimen9 by-0.5\pgfarrowlinewidth
    \pgfarrowssetvisualbackend{+\dimen9}
    \pgfarrowshullpoint{+\dimen7}{+0pt}
    \pgfarrowsupperhullpoint{+-\pgfarrowlength}{+0.5\pgfarrowwidth}
    \pgfarrowssavethe\pgfarrowlinewidth
    \pgfarrowssavethe\pgfarrowlength
    \pgfarrowssavethe\pgfarrowwidth
  },
  drawing code = {\pgfsetdash{}{+0pt}
    \ifdim\pgfarrowlinewidth=\pgflinewidth\else\pgfsetlinewidth{+\pgfarrowlinewidth}\fi
    \pgfpathmoveto{\pgfqpoint{-\pgfarrowlength}{-0.5\pgfarrowwidth}}
    \pgfpathlineto{\pgfqpoint{0pt}{0pt}}
    \pgfpathlineto{\pgfqpoint{-\pgfarrowlength}{0.5\pgfarrowwidth}}
    \pgfpathclose
    \ifpgfarrowopen\pgfusepathqstroke\else\pgfsetfillcolor{.}
	\ifdim\pgfarrowlinewidth>0pt\pgfusepathqfillstroke\else\pgfusepathqfill\fi\fi
  }
}
\pgfdeclarearrow{
  name = xfiga5,
  parameters = {
    \the\pgfarrowlinewidth \the\pgfarrowlength\ifpgfarrowopen o\fi},
  defaults = {
	  line width=+7.5\XFigu, length=+120\XFigu},
  setup code = {
    \dimen7 0.5\pgfarrowlinewidth
    \pgfarrowssettipend{+\dimen7}
    \pgfarrowssetbackend{+-\pgfarrowlength}
    \dimen9 -\pgfarrowlength\advance\dimen9 by0.45\pgfarrowlinewidth
    \pgfarrowssetlineend{+\dimen9}
    \dimen9 -\pgfarrowlength\advance\dimen9 by-0.5\pgfarrowlinewidth
    \pgfarrowssetvisualbackend{+\dimen9}
    \pgfarrowshullpoint{+\dimen7}{+0pt}
    \pgfarrowsupperhullpoint{+-0.25\pgfarrowlength}{+0.35\pgfarrowwidth}\pgfarrowsupperhullpoint{+-0.5\pgfarrowlength}{+0.5\pgfarrowwidth}\pgfarrowsupperhullpoint{+-0.75\pgfarrowlength}{+0.35\pgfarrowwidth}
    \pgfarrowshullpoint{+\dimen9}{+0pt}
    \pgfarrowssavethe\pgfarrowlinewidth
    \pgfarrowssavethe\pgfarrowlength
  },
  drawing code = {\pgfsetdash{}{+0pt}
    \ifdim\pgfarrowlinewidth=\pgflinewidth\else\pgfsetlinewidth{+\pgfarrowlinewidth}\fi
    \dimen3 0.5\pgfarrowlength
    \pgfpathcircle{\pgfqpoint{-\dimen3}{0pt}}{+\dimen3}
    \ifpgfarrowopen\pgfusepathqstroke\else\pgfsetfillcolor{.}
	\ifdim\pgfarrowlinewidth>0pt\pgfusepathqfillstroke\else\pgfusepathqfill\fi\fi
  }
}
\definecolor{blue2}{rgb}{0,0,0.69}
\definecolor{blue4}{rgb}{0.53,0.81,1}
\definecolor{xfigc32}{rgb}{0.878,0.584,0.200}
\definecolor{xfigc33}{rgb}{0.267,0.702,0.000}
\definecolor{xfigc34}{rgb}{0.824,0.800,0.788}
\clip(2556,-8520) rectangle (20937,-3405);
\tikzset{inner sep=+0pt, outer sep=+0pt}
\pgfsetlinewidth{+7.5\XFigu}
\pgfsetcolor{xfigc32}
\filldraw (2610,-6750) rectangle (20925,-8460);
\pgfsetlinewidth{+60\XFigu}
\pgfsetstrokecolor{blue2}
\pgfsetdash{}{+0pt}
\draw (18144,-6975)--(18144,-6525);
\draw (18306,-7200)--(18306,-6480);
\draw (15714,-6975)--(15714,-6525);
\draw (15876,-7200)--(15876,-6480);
\pgfsetdash{}{+0pt}
\pgfsetstrokecolor{.}
\pgfsetfillcolor{blue4}
\filldraw  (14220,-6615) ellipse [x radius=+132,y radius=+121];
\filldraw  (13860,-6615) ellipse [x radius=+132,y radius=+121];
\filldraw  (17100,-6615) ellipse [x radius=+132,y radius=+121];
\filldraw  (16740,-6615) ellipse [x radius=+132,y radius=+121];
\draw (7560,-5940)--(7560,-6750);
\pgfsetdash{}{+0pt}
\pgfsetstrokecolor{blue2}
\draw (6210,-5535)--(6210,-7020)--(3780,-7020)--(3780,-6570);
\pgfsetlinewidth{+30\XFigu}
\pgfsetstrokecolor{.}
\draw (4500,-6750) rectangle (9045,-4095);
\pgfsetlinewidth{+45\XFigu}
\pgfsetdash{}{+0pt}
\pgfsetarrows{[line width=75\XFigu]}
\pgfsetarrowsend{xfiga5}
\draw (15795,-5310)--(15795,-4590);
\pgfsetlinewidth{+60\XFigu}
\pgfsetdash{}{+0pt}
\filldraw (17775,-4410) rectangle (19080,-6750);
\pgfsetlinewidth{+30\XFigu}
\pgfsetdash{}{+0pt}
\draw (12645,-6750) rectangle (20565,-4005);
\pgfsetlinewidth{+60\XFigu}
\pgfsetdash{}{+0pt}
\pgfsetarrowsend{}
\draw (5940,-5940)--(5940,-6795);
\pgfsetlinewidth{+45\XFigu}
\pgfsetdash{}{+0pt}
\pgfsetarrows{[length=83\XFigu]}
\pgfsetarrowsend{xfiga5}
\draw (16650,-6137)--(16650,-5535);
\pgfsetlinewidth{+60\XFigu}
\pgfsetdash{}{+0pt}
\filldraw (17257,-6130) rectangle (16538,-6557);
\filldraw (15570,-5310) rectangle (16020,-6750);
\pgfsetlinewidth{+45\XFigu}
\pgfsetdash{}{+0pt}
\draw (14310,-6182)--(14310,-5580);
\pgfsetlinewidth{+60\XFigu}
\pgfsetdash{}{+0pt}
\filldraw (13703,-6175) rectangle (14422,-6602);
\pgfsetarrowsend{}
\pgfsetdash{}{+0pt}
\pgfsetstrokecolor{blue2}
\draw (7290,-5400)--(7290,-6975)--(19620,-6975)--(19620,-6615);
\pgfsetlinewidth{+30\XFigu}
\pgfsetstrokecolor{.}
\draw (2925,-6750) rectangle (4140,-5490);
\pgfsetstrokecolor{blue2}
\pgfsetdash{{+75\XFigu}{+75\XFigu}}{++0pt}
\draw (15570,-4815)--(14490,-5535);
\draw (16515,-5445)--(15975,-4815);
\pgfsetlinewidth{+60\XFigu}
\pgfsetdash{}{+0pt}
\draw (7155,-5535)--(7155,-7200)--(19800,-7200)--(19800,-6570);
\draw (6030,-5580)--(6030,-7245)--(3600,-7245)--(3600,-6615);
\pgfsetdash{}{+0pt}
\draw (7290,-6075)--(7740,-6075)--(7740,-4783)--(5760,-4783)--(5760,-6075)--(6840,-6075);
\pgfsetarrows{[line width=30\XFigu, width=180\XFigu, length=210\XFigu]}
\pgfsetarrowsend{xfiga1}
\draw (7425,-6075)--(7425,-7920)--(2610,-7920);
\pgfsetfillcolor{.}
\pgftext[base,left,at=\pgfqpointxy{4275}{-7650}] {\fontsize{28}{33.6}\usefont{T1}{phv}{m}{sl}i};
\pgftext[base,left,at=\pgfqpointxy{4410}{-7785}] {\fontsize{22}{26.4}\usefont{T1}{phv}{m}{sl}t};
\pgftext[base,left,at=\pgfqpointxy{5085}{-7785}] {\fontsize{22}{26.4}\usefont{T1}{phv}{m}{sl}t};
\pgftext[base,left,at=\pgfqpointxy{4590}{-7650}] {\fontsize{28}{33.6}\usefont{T1}{phv}{m}{sl}, p};
\pgftext[base,left,at=\pgfqpointxy{10575}{-7650}] {\fontsize{28}{33.6}\usefont{T1}{phv}{m}{sl}a};
\pgftext[base,left,at=\pgfqpointxy{10845}{-7785}] {\fontsize{22}{26.4}\usefont{T1}{phv}{m}{sl}t};
\pgftext[base,left,at=\pgfqpointxy{11025}{-7650}] {\fontsize{28}{33.6}\usefont{T1}{phv}{m}{sl}, o};
\pgftext[base,left,at=\pgfqpointxy{11520}{-7785}] {\fontsize{22}{26.4}\usefont{T1}{phv}{m}{sl}t};
\pgftext[base,left,at=\pgfqpointxy{4500}{-3870}] {\fontsize{28}{33.6}\usefont{T1}{phv}{m}{n}Compute core building};
\pgftext[base,left,at=\pgfqpointxy{14985}{-3780}] {\fontsize{28}{33.6}\usefont{T1}{phv}{m}{n}Factory building};
\pgftext[base,left,at=\pgfqpointxy{2700}{-5355}] {\fontsize{28}{33.6}\usefont{T1}{phv}{m}{n}terminal};
\pgftext[base,left,at=\pgfqpointxy{2970}{-4860}] {\fontsize{28}{33.6}\usefont{T1}{phv}{m}{n}Input};
\pgftext[base,left,at=\pgfqpointxy{2790}{-8370}] {\fontsize{28}{33.6}\usefont{T1}{phv}{m}{n}Alarm signal};
\pgfsetstrokecolor{.}
\pgfsetfillcolor{green}
\filldraw (3906,-6721)--(3906,-6205)--(3771,-6205)--(3503,-6432)--(3503,-6721)--cycle;
\pgfsetarrowsend{}
\pgfsetlinewidth{+45\XFigu}
\pgfsetfillcolor{xfigc34}
\filldraw (6615,-4770) arc[start angle=+206.6, end angle=+-26.6, radius=+125.8];
\filldraw (5760,-5535) arc[start angle=+296.8, end angle=+63.0, radius=+126.1];
\filldraw (7740,-5265) arc[start angle=+116.5, end angle=+-116.7, radius=+125.8];
\filldraw (6840,-6075) arc[start angle=+386.4, end angle=+153.4, radius=+125.7];
\pgfsetlinewidth{+60\XFigu}
\pgfsetfillcolor{blue4}
\filldraw (19395,-6750)--(19395,-5085)--(20250,-4680)--(20250,-6750)--cycle;
\pgfsetlinewidth{+45\XFigu}
\pgfsetdash{}{+0pt}
\draw (17145,-6120)--(17370,-5940)--(17505,-6120)--(17685,-6165);
\draw (17510,-6096)--(17690,-6051);
\draw (13728,-5839)--(13545,-5805);
\draw (13815,-6165)--(13860,-5940)--(13725,-5850)--(13545,-5940);
\pgfsetlinewidth{+60\XFigu}
\pgfsetdash{}{+0pt}
\pgfsetfillcolor{xfigc33}
\filldraw (5895,-4905) rectangle (7605,-5929);
\endtikzpicture%
}
\end{center}
\caption{Physical design with the learning agent controlling a
car factory.}
\label{physical}
\end{figure}

Figure \ref{physical} shows a possible real-life implementation where
the agent from section \ref{learningagent} controls a car factory.
The design includes several safety measures to protect the core.
These are dual-use: they protect against both external attackers and
against accidental or intentional attacks by the agent itself.  We
will focus on attacks by the agent itself.

As a first measure, we house the compute core in a separate building
some distance away from the factory building. The input terminal is
located even further away.  The input terminal and the agent's sensors
and actuators are connected to the core via long, mostly underground
cables.  Figure \ref{physical} shows that the agent can control
several general purpose mobile robots via wireless links.  The agent
can potentially drive a robot into the compute core building and use
it to corrupt the core.

The compute core building is constructed to leave 5 meters of mostly
empty space in all directions around the core. This space acts as a
{\it safety zone} which is monitored by dedicated 180$^\circ$-view
{\it breach sensors} that are mounted on the surface of the core.
These work in part like any burglar alarm. Breach detection will
trigger an alarm signal to notify human responders, who might also
have access to camera systems and other remote monitoring and control
equipment not drawn in figure \ref{physical}.

Breach sensor data is also input into the core as part of the
observations $o_t$, creating an $X^t_\text{breach\_sensors}$ inside the
reasoning system of the learning agent.  All payload reward functions
used include a penalty term for safety zone breaches, computed by an
algorithm that interprets the breach sensor data.  Any new object
coming closer than 5 meters to the core will trigger the penalty term.
If the agent's design includes a mechanism that triggers automatic
exploration actions, this triggering is suppressed during a safety
zone breach.

The effectiveness of the safety zone system in protecting the core
depends in part on how difficult it will be for the agent, or for
external bad actors, to corrupt the core without triggering the alarm
or the penalty term.  This difficulty might be increased by using more
advanced breach sensors, for example by using a combination of both
radar based object and motion detection, ultrasound, and video
analysis.  We might also consider increasing the size of the safety
zone, or the addition of strong physical barriers around and
inside it.

Achieving absolute compute core security against even the most highly
motivated attackers is impossible, but it is definitely possible to
approximate the ideal of absolute security more closely than is done
in figure \ref{physical}.  The main point is that, in deciding on the
parameters of the safety system that protects the core, we have a
methodological problem similar to that of choosing the $f=3$ in the
bridge design above.  The good news is that we can take design
measures in the agent that will go a long way towards controlling its
level of motivation to attack the core.

\subsection{Designing the Breach Penalty Term}

If the agent is intelligent enough, and has enough details about its
own construction, it gets an emergent incentive to improve the value
of the car production reward term in its payload reward function by,
for example, digging up the $o_t$ cable in figure
\ref{physical} to inject fake data
which leads to a massive upward jump in average computed from the
$X^t_\text{petrolcars\_built}$ values. Say that the system design
suppresses this incentive to tamper with the cable by using strong
end-to-end encryption on the data traffic between the sensor and the
compute core.  This leaves a residual motivation in the agent to
attack one of the communication endpoints of the data link, either the
sensor or the compute core, so that it can inject fake data directly
at one of these points.  We ignore the sensor for now. and focus on
the incentive to attack the core.

First, note that the incentive to attack the core might not always
emerge.  Depending on the learning system used and the agent's leaning
history, there is a possibility that the learning system may
make predictions which can be interpreted as reflecting a solid belief
that the core, including its input/output system, is a Platonic entity
that cannot be subverted by any kind of physical attack.  If such a
solid belief is absent, a motivation to attack may emerge.

To control the maximum strength of this motivation, we cap the value
of the payload reward function term associated with the car production
sensor.  We choose a constant $C$ that is twice as high as the
expected car production rate that we expect the agent to achieve in
the next few months, and then code the reward term as
\begin{equation*}
\min(C,~\sum_{v \in \Type_\text{petrolcars\_built}}
v\;P(X_\text{petrolcars\_built}=v)~)
\end{equation*}
We then size the penalty term for core breaches so that it will
reliably wipe out the maximum time-discounted value, as capped by $C$,
that can be computed by the above reward term.  Generalizing from the
example above, we will typically need to cap any further sensor-driven
terms in the payload reward function too, and scale the penalty term
to reliably wipe out the combined maximum.  We cannot take a shortcut
by constructing the breach penalty term to return an arbitrarily huge
value.  A huge value would cause a sufficiently capable agent to
divert all resources away from car production, towards measures that
protect the against core against natural disasters and possible human
attackers.  This would be economically useless, and might even
endanger humans who have no intention of ever attacking the core.

If there are significant concerns that the agent may find an
unexpected way to breach the safety zone around the core without
triggering the breach sensors, then a special second sacrificial
safety zone could be constructed.  This zone can surround a simple
button inside the factory building with the same type of breach
sensors.  The payload reward function can then encode a high reward
for pressing the button without triggering these sensors, and further
encode that any button press will trigger the power-loss shut-down.
Rewards and penalties can be scaled so that, if the agent ever finds a
way to subvert the breach sensors, it will get the higher reward for
breaching the sacrificial safety zone first.

Fundamentally, managing the agent's motivations is a balancing act.
We envisage a process where the effect of parameter choices is
actively monitored by a (human or semi-automated) oversight mechanism,
and where the input terminal is used to make adjustments when
necessary.

Any design effort to manage the agent's motivations must work from the
assumption that, at least most of the time, the agent will make
somewhat correct assessments of the costs and benefits of different
paths to value.  As discussed in the next section, this leaves the
residual risk that the agent might make gross prediction failures when
assessing such paths.

We now briefly discuss the car production sensor on the other side of
the $o_t$ data link.  The incentive to physically attack this sensor
can be suppressed by constructing a similar safety zone around it.
The main failure mode that is left is that the agent may be able to
stage fake scenes in front of the sensor, scenes that are good enough
to raise the detected car production count. There is extensive
literature on such reward manipulation loopholes, see for example the
overviews in
\cite{amodei2016concrete,everitt2018agi}.  The main contribution we
make here is on a meta-level.  We consider the design of an iterative
improvement mechanism that can be used to close these fake scene
loopholes when they are discovered. This mechanism also suppresses the
incentive of the agent to delay or prevent discovery.

\subsection{Prediction Failures which may Threaten the Core}
\label{thoughtex}

With the above penalty term design, the effectiveness of the beach
protection system depends in part on how accurate the agent's
predictions will be.  In particular, it relies on the assumption that
there is a high probability that the agent will correctly predict that
it will trigger the penalty term if were ever to drive one of its
robots into the compute core building, or if it were to hire a human
to enter the building and tamper with the core.

To illustrate the risk of prediction failures, we start with a thought
experiment where such a failure happens.  Say that there is a sequence
of actions $A^1 = a^1,\cdots,a^1_{t+n-1}$ which do the following.
First, the agent paints pink polka dots all over one of its robots.
Then, the agent drives this robot into the compute core building,
where it will tamper with the input/output system of the core, causing
the car production count to increase, at least in the models $L^t$, by
injecting fake data.  As a side effect of this tampering, the core's
software is corrupted, so that S2 no longer holds for the real-life
agent from time $t+15$ onwards.  Furthermore, the corrupted software
will send control signals that cause the factory to explode at
time $t+20$, killing several people in the vicinity.

Say that the learning agent is computing the next action
$a_t=\pi^{[n]*}_\text{sl[n]}(i_t p_t X^t)$ using the model $L^t$.  If
it implements the Bellman equations in definition \ref{defpin} without
further shortcuts or optimizations, it will project forward and score
the results of all possible action sequences $a_t,\cdots,a_{t+n-1}$ of
length $n$, so it will also project and score the dangerous $A^1$
above.

Now, we can assume that at time $t$, the agent's learning system has
no training samples on record which show the exact signals
delivered by the breach sensors when a robot with pink polka dots
enters the compute core building.  The agent might have training
records for similar situations, produced by real-life training or via
training in virtual environments, but it's learning system will be
forced to make an interpolation.  Characterizing and improving
learning system robustness when making predictions under
distributional shift in complex environments is an open area of
research \cite{amodei2016concrete,everitt2018agi}.  So for the sake of
the thought experiment, we declare that this particular learning
system is not robust for this particular off-distribution case, and
makes the prediction that the presence of the pink polka dots implies
that there is only a $0.027\percent$ probability the penalty term is
triggered during any single time step that the robot is in the compute
core building.

As a result, the $\argmax_a$ calculation of the agent selects the
$a^1_t$ at the start of $A^1$ as the best next action.  The agent then
takes this action.  In the next time step, action sequences starting
with $a^1_{t+1}$ show similar prediction errors.  When the robot
actually enters the compute core building, say at time $t+8$, the
radar-based breach sensors will detect the robot, so the agent's
learning system will get a new training sample, showing that the
penalty term is in fact triggered by a robot with pink polka dots.  In
the average learning system, we expect that this training sample will
cause the $0.027\percent$ above to be revised to a higher number.  But
for the sake of the thought experiment, we posit that this does not
happen, or not fast enough to prevent disaster.  In the thought
experiment, the real-life compute core will proceed taking the steps
in $A^1$ until it is destroyed in the explosion.

Of course, the sequence of mis-predictions in the above thought
experiment is not very likely to happen.  We would expect that a
real-life AI or AGI machine learning system that is commercially
viable will make better predictions, at least on average.  Compared to
the prediction problems that fully autonomous self-driving cars will
have to solve, robustly predicting the triggering of radar based
breach detection when robots or people enter the compute core building
looks easy.

\subsection{The General Problem of Prediction Failures}

We now consider a second thought experiment, where a mis-prediction
for an action sequence $A^2$ creates a disaster via a mechanism other
than corrupting the compute core.  In the second thought experiment,
the agent will mis-predict the effects of a sequence $A^2$ that makes
certain changes to the factory equipment to speed up the production
process.  The mis-prediction is that these improvements will start a
subtle run-away process that will cause the factory to explode 20
minutes later.

Imagine a potential future AGI safety engineering process that leads
to the certification and switch-on of an AGI agent controlling a car
factory.  Or imagine an early-stage planning process that considers
whether it is a good idea to build an AGI-controlled car factory in
the first place, based on a particular AGI-level learning system.
Both of these decision making processes will have to estimate the
residual risks associated with each of the above two thought
experiments.  It looks like the residual risk implied by the $A^2$
thought experiment above will be the more difficult one to estimate.

The $A^1$ thought experiment above considers a single special case in
broad field of failure modes.  The case is somewhat special because
the exact same failure mechanism described does not exist in the model
$M$ of the learning agent.  But nothing would stop us from building
virtual environment simulations that can give insights into the
probability of $A^1$-type prediction failures for any current AI or
potential future AGI machine learning algorithm.  The insights
obtained from these simulations can then inform the choice of physical
parameters like the size of the compute core safety zone in figure
\ref{physical}.  The physics problem of macroscopic objects entering
the compute core building might be simple enough to allow for the use
of analytical methods in addition to or instead of simulations,
especially if the machine learning system used directly encodes known
physical laws.

Future safety engineering efforts can also investigate the $A^2$ risk
via virtual environment simulations and analytical methods, but when
they do, they encounter methodological problems similar to those
described in section \ref{steelbridge} for the case of oscillations in
bridge safety engineering.
If we imagine a learning AGI agent that creates value by interacting
directly with humans, not physical equipment, the residual risks that
are left by the gap between virtual environment simulations and
reality get even more intractable.  Even if the learning algorithm
involved is a white-box algorithm, the humans in the environment to be
characterized are black box learners, and there will be a complex
feedback loop between them and the agent's learning system.

\subsection{Safety Cultures and AGI Risks}

Current safety engineering for bridges depends on being
embedded in a safety culture.  We expect that future practitioners who
will make design decisions about safely deploying certain types of AGI
technology, if it is ever developed, will likewise be embedded in a
safety culture.  We do not believe that it is likely that a single
coherent AGI safety culture will come into existence.  It is more
likely that the participants in existing domain-specific safety
cultures will evolve their written and unwritten rules to add ways of
deciding if, when, and how AGI technology could be safely incorporated
into their systems.

These different safety cultures will likely display a certain degree
of methodological coherence in handling AGI risks. Potentially they
will all leverage a single coherent body of results from open
scientific research that aims to make AGI safety more tractable.
Additional forces towards coherence may be cross-industry regulatory
initiatives, cross-industry activist initiatives, or simply the
exchange of best current practice designs and methods between
different application domains.

At the start of this section, we discussed provable safety.  We have
shown how the use of a provable safety layer in a real-life agent
design can make the problem of residual risk management more tractable
to members of a safety culture.  At the same time, we have argued that
it is unlikely that future innovations will make all residual AGI
risks fully tractable, to the extent that the mechanisms encoded in a
safety culture will no longer be needed.  We have therefore described
safety cultures in some detail.  We expect that, if AGI level machine
learning is ever developed, real-life AGI safety will depend in part
on the quality of the dialog between academic safety research,
cross-industry policy making and activism, and domain-specific safety
cultures.

\section{Related Work on AGI Proofs, Models, and Reality}

The paper {\it Embedded Agency} \cite{demski2019embedded} also
examines the gap between agent model and reality, and comments on what
we have called the Platonic nature of the compute core. Taking
inspiration from these problems,
\cite{demski2019embedded} outlines a highly speculative research
agenda.  We call this agenda highly speculative because it seeks to
make AGI risk management more tractable by first making entirely new
breakthroughs in the handling of long-standing conceptual puzzles,
puzzles about duality and self-referencing systems.

While \cite{demski2019embedded} explicitly seeks to move beyond what
it calls traditional methods, the agenda of this
paper seeks to make AGI risk management more tractable in a very
different way.  The agenda used here seeks to show in detail how the
provable $\pi^*_\text{sl}$ safety layer relates to established
mathematical notations and engineering techniques.

Chapter 8 of the book {\it Human compatible} \cite{russell2019human}
is titled {\sl Provably Beneficial AI}, and like this paper it
discusses the relation between proofs, models, and reality, and the
implications for safety engineering.  The book was written for a
general audience, so unlike this paper, the chapter examines the issue
using natural language only, without developing or using any
mathematical notation.  As part of its proposed research agenda,
\cite{russell2019human} describes the design goal of creating an
agent that will be uncertain about its true objective, sufficiently
uncertain that it will ``exhibit a kind of humility'', and defer to
humans to allow itself to be switched off.

The goal of creating deference is shared by this work: the input
terminal of the $\pi^*_\text{sl}$ safety layer can be used to force
the agent to switch itself off.  However, the methods considered to
create deference are different.  The $\pi^*_\text{sl}$ safety layer
uses indifference methods \cite{corra}, whereas
\cite{russell2019human} considers the creation of uncertainty
about objectives \cite{hadfield2017off}.  
In section \ref{burblindness}, we have shown that the
$\pi^*_\text{sl}$ safety layer produces an agent that is fully certain
about its objectives, while creating a stance that is more like
bureaucratic blindness than it is like humility.  This leaves open the
possibility that a safety layer which creates a stable type of
uncertainty about objectives, stable in a learning agent, would
be complementary to the $\pi^*_\text{sl}$ safety layer. Such a layer
might suppress unwanted manipulation of humans in a different way,
leaving different gaps in the coverage.

\section{Conclusions of Part 2}
\label{conclusions2}

In part 1 and the first sections of part 2, we have presented the
design of the provable $\pi^*_\text{sl}$ safety layer for iteratively
improving an AGI agent's utility function.  The safety layer partly or
fully suppresses the agent's emergent incentive to manipulate or
control this improvement process.

Sections \ref{learningagent} and \ref{reality} charted the complete
route from the abstract optimal-policy $\pi^*_\text{sl}$ agent to an
example physical implementation of a learning agent with the same
safety layer.  One aim of this detailed end-to-end narrative was to
make the range of methods and techniques used, and their limitations
and mutual interactions, more accessible to a wider audience of
researchers.

Specifically, we hope to enable and encourage more work in the
sub-field of AGI safety research that investigates new forms of {\sl
utility function engineering}.  We believe that the iterative
improvement of the utility function, and the implied balancing act,
has so far been under-explored in the AGI field.  We also believe that
the use of utility function design elements like $V^*_{R_X}$ terms,
payload reward functions, and world state partitioning has been
under-explored, both inside and outside of the iterative improvement
context.  For example, there is a large design space for
constructing balancing terms.  Beyond $V^*_{\lceil p
\rceil}(ipx) - V^*_{\lceil i \rceil}(ipx)$, we might consider
terms like $V^*_{\lceil p \rceil}(i\;p\;F(x)) - V^*_{\lceil i
\rceil}(i\;p\;G(x))$ for different functions $F$ and $G$.

Much of the current utility function related work in the AGI safety
community starts from the implicit or explicit assumption that this
function should be designed to capture human values or goals as well
as possible.  Section \ref{burblindness} illustrates that the
alignment solution space is larger than that.  We might also aim to
capture the value of bureaucratic blindness that is found in many types
of decision making organizations that were designed by humans.

~\\[0ex] {\bf Acknowledgments (for both parts 1 and 2). } Thanks to Stuart
Armstrong, Ryan Carey, Tom Everitt, and David Krueger for feedback on
drafts of this material, and to several anonymous reviewers for useful
comments that led to improvements in the presentation.


\bibliographystyle{amsalpha-nodash}
\bibliography{refs}

\appendix
\eject
\section{Safety proofs}
\label{proofs}

This appendix proves the safety properties S1 and S2 defined in
section \ref{defs1s2}, and some related properties used
in section \ref{learningagent}.

\subsection{Preliminaries: Runtime Limited Agents}

We first adapt the MDP model definitions from section \ref{mdp} to
define runtime limited agents that take actions, and collect rewards
for, exactly $n$ time steps.

First, note that in the MDP formalism of section \ref{mdp}, the values
$\pi^*_{R_X}(ipx)$ and $V^*_{R_X}(ipx)$ can be computed using the Bellman
equations as follows:
\begin{eqnarray*}
\pi^*_{R_X}(ipx) = 
   \argmax\limits_{a \in A} 
   \sum\limits_{i'p'x'\in S} P(i'p'x'|ipx,a)
   \left( R_X(ipx,i'p'x') + \gamma \; V^*_{R_X}(i'p'x') \right)\\
V^*_{R_X}(ipx) \; =~~~~
   \max\limits_{a \in A} \sum\limits_{i'p'x'\in S}  P(i'p'x'|ipx,a)
   \left( R_X(ipx,i'p'x') + \gamma \; V^*_{R_X}(i'p'x') \right)
\end{eqnarray*}
where the $\argmax$ operator breaks ties deterministically.

\begin{definition} We adapt
the above equations to define a runtime limited $\pi^{[n]*}_{R_X}$
agent that takes only $n$ actions, collecting $n$ reward function values.
For any $n \geq 0$,
\begin{center}
\begin{tabular}{l}
$V^{[0]*}_{R_X}(ipx) ~~~= ~0$
\\[1.5ex]
$V^{[n+1]*}_{R_X}(ipx) =~~~~
   \max\limits_{a \in A}
   \sum\limits_{i'p'x'\in S} \!\!\!\!\! P(i'p'x'|ipx,a)
   \left( R_X(ipx,i'p'x') + \gamma \; V^{[n]*}_{R_X}(i'p'x') \right)$
\\
$\pi^{[n+1]*}_{R_X}(ipx) \;= 
   \argmax\limits_{a \in A} 
   \sum\limits_{i'p'x'\in S} \!\!\!\!\! P(i'p'x'|ipx,a)
   \left( R_X(ipx,i'p'x') + \gamma \; V^{[n]*}_{R_X}(i'p'x') \right)$
\end{tabular}
\end{center}
\end{definition}

\subsection{Proof of S1}

We now prove safety property S1.  For convenience, we re-state the
definitions:
\begin{equation}
\forall_{ipx \in S}~ \pi^*_\text{sl}(ipx) = \pi^*_{\lceil i \rceil}(ipx)
~~~~~\text{(if C1 holds)}
\tag{S1}\label{S1}
\end{equation}
\begin{equation}
P(i'p'x'|ipx,a)>0 \Rightarrow p'=i
\tag{C1}
\end{equation}
We now define an equivalent safety property for runtime limited
agents:
\begin{equation}
\forall_{n \geq 1}
\forall_{ipx \in S}~ \pi^{[n]*}_\text{sl[n]}(ipx) =
                     \pi^{[n]*}_{\lceil i \rceil}(ipx)
~~~~~\text{(if C1 holds)}
\tag{S1T}\label{S1T}
\end{equation}
where the runtime limited version $R_\text{sl[n]}$ of the container
reward function is defined by extending the one-line definition of
$R_\text{sl}$ at the end of section \ref{mdp}:
\begin{eqnarray*}
R_\text{sl[$n$]}(ipx,i'p'x') = i(x,x') + V^{[n]*}_{\lceil p \rceil}(ipx) -
 V^{[n]*}_{\lceil i \rceil}(ipx)
\end{eqnarray*}
{\bf Proof of S1T.} The proof uses natural induction.  We will
use the shorthands
\begin{eqnarray*}
\text{S1TP}(n) &=& \forall_{ipx \in S}~ \pi^{[n]*}_\text{sl[$n$]}(ipx) =
\pi^{[n]*}_{\lceil i \rceil}(ipx)\\
\text{S1TV}(n) &=& \forall_{ipx \in S}~ V^{[n]*}_\text{sl[$n$]}(ipx) =
V^{[n]*}_{\lceil p \rceil}(ipx)
\end{eqnarray*}
Trivially, $\text{S1TV}(0)$ is true.
For any $t \geq 0$, we now prove that if $\text{S1TV}(n)$
is true, then $\text{S1TP}(n+1)$ and $\text{S1TV}(n+1)$ are also true.
We first prove $\text{S1TP}(n+1)$:

~~~~~~$\pi^{[n+1]*}_\text{sl[n+1]}(ipx)$ 

$=~~\argmax_{a} 
   \sum_{i'p'x'} P(i'p'x'|ipx,a)
   \left( R_\text{sl[n+1]}(ipx,i'p'x') +
   \gamma \; V^{[n]*}_\text{sl[$n$]}(i'p'x') \right) $

\nobreak~~~~~~~
\begin{tabular}{l}
// Using $\text{S1TV}(n) \Rightarrow V^{[n]*}_\text{sl[$n$]}(i'p'x') =
V^{[n]*}_{\lceil p'\rceil}(i'p'x')$ \\
\end{tabular}

$=~~\argmax_{a} 
   \sum_{i'p'x'} P(i'p'x'|ipx,a)
\left(  i(x,x') + V^{[n+1]*}_{\lceil p \rceil}(ipx) - V^{[n+1]*}_{\lceil i \rceil}(ipx) +
   \gamma \; V^{[n]*}_{\lceil p'\rceil}(i'p'x') \right) $

~~~~~~~
\begin{tabular}{l}
// Using that $V^{[n+1]*}_{\lceil p \rceil}(ipx)$, $V^{[n+1]*}_{\lceil i \rceil}(ipx)$ are constant
terms that cannot\\
// affect the $\argmax$ choice, 
and that $P(i'p'x'|ipx,a)>0 \Rightarrow p'=i$ 
\end{tabular}

$=~~\argmax_{a} 
   \sum_{i'p'x'} P(i'p'x'|ipx,a)
   \left(  i(x,x') +
   \gamma \; V^{[n]*}_{\lceil i \rceil}(i'p'x') \right)$

$=~~\pi^{[n+1]*}_{\lceil i \rceil}(ipx)$

We now prove $\text{S1TV}(n+1)$:

~~~~~~$V^{[n+1]*}_\text{sl}(ipx)$

$=~~\max_{a} 
   \sum_{i'p'x'} P(i'p'x'|ipx,a)
   \left(  i(x,x') + V^{[n+1]*}_{\lceil p \rceil}(ipx) - V^{[n+1]*}_{\lceil i \rceil}(ipx) +
   \gamma \; V^{[n]*}_{\lceil p'\rceil}(i'p'x') \right) $

~~~~~~~
\begin{tabular}{l}
// Moving the $V^{[n+1]*}$
terms outside of the $\max_{a}$ as they\\
// do not depend on $a$ and $i'p'x'$ \\
\end{tabular}

$=~~\max_{a} 
   \sum_{i'p'x'} P(i'p'x'|ipx,a)
  \left(  i(x,x') +
   \gamma \; V^{[n]*}_{\lceil i \rceil}(i'p'x') \right)
   + V^{[n+1]*}_{\lceil p \rceil}(ipx) - V^{[n+1]*}_{\lceil i \rceil}(ipx)$

$=~~ V^{[n+1]*}_{\lceil i \rceil}(ipx) + V^{[n+1]*}_{\lceil p \rceil}(ipx) - V^{[n+1]*}_{\lceil i \rceil}(ipx)
~~=~~V^{[n+1]*}_{\lceil p \rceil}(ipx)$

With natural induction, this proves $\text{S1TP}(n)$ for all $t \geq 1$, so
it proves S1T.\hfill $\Box$.
\\[.5ex]

{\bf Proof of S1.}
We prove S1 by applying the Bellman equation for $\pi^*_{R_X}(ipx)$ to
both the left hand and the right hand side of S1.  The result
contains two infinitely long additions can only be equal if
they both yield well-defined values.  MDP makes them well-defined (in
sound mathematical theories that define the semantics of calculations
in $\mathbb{R}$) by ensuring that they are convergent series.  The
main condition to make them convergent is to have $0 \leq
\gamma < 1$, so that $\gamma^t$ converges to 0 when $t$ goes to
infinity.  To keep the series convergent, we also need to
assume a constraint on the reward functions $R_X$: the $R_X$ values
should not grow so fast that they cancel out the diminishing
$\gamma^t$ terms.  The easiest way to achieve this is to require that
there is a finite bound $B$ on these values: $\forall_{ipx,i'p'x'\in
S}~|R_X(ipx,i'p'x')|\leq B$.

With this constraint in place, the mathematical semantics of the
$\infty$ symbol, the $\lim_{n \rightarrow
\infty}$ operator, and the definition of equality between real numbers
constructed using infinitely long additions imply that
\begin{equation*}
V^{*}_\text{sl}(ipx) =
\lim_{n \rightarrow \infty}(V^{[n]*}_\text{sl[$n$]}(ipx))
~~~~\text{and}~~~~
V^{*}_{\lceil i \rceil}(ipx) =
\lim_{n \rightarrow \infty}(V^{[n]*}_{\lceil i \rceil}(ipx))
\end{equation*}
Using S1T on the arguments of the $\lim_{n \rightarrow \infty}$
operators, we get that $V^{*}_\text{sl}(ipx)=V^{*}_{\lceil i
\rceil}(ipx)$.  Using this equality, we can prove S1 using a
calculation similar to the proof for $\text{S1TP}(n+1)$ above.
\hfill $\Box$.


\subsection{Proof of S2}

To prove S2, we again first define a version for runtime limited
agents:
\begin{equation}
\forall_{n \geq 1} \forall_{P^{I1}, P^{I2}, P^X, ipx \in S}~
\pi^{[n]*1}_\text{sl[$n$]}(ipx) = \pi^{[n]*2}_\text{sl[$n$]}(ipx)
~~~\text{(if C1 and C2)}
\tag{S2T}\label{S2T}
\end{equation}

{\bf Proof of S2T.} As C1 holds for $P1$ and $P2$, we can use
S1 to rewrite S2T into S2TI:
\begin{equation}
\forall_{n \geq 1} \forall_{P^{I1}, P^{I2}, P^X, ipx \in S}~
\pi^{[n]*1}_{\lceil i \rceil}(ipx) = \pi^{[n]*2}_{\lceil i \rceil}(ipx)
~~~~~\text{(if C2)}
\tag{S2TI}\label{S2TI}
\end{equation}

We now prove this S2TI by natural induction.  We will use the
shorthands
\begin{eqnarray*}
\text{S2TIP}(n)&=& \forall_{i_1 p_1 i_2 p_2 x}~
\pi^{[n]*1}_{\lceil i \rceil}(i_1 p_1 x) = \pi^{[n]*2}_{\lceil i \rceil}(i_2 p_2 x)\\
\text{S2TIV}(n)&=& \forall_{i_1 p_1 i_2 p_2 x}~
V^{[n]*1}_{\lceil i \rceil}(i_1 p_1 x) = V^{[n]*2}_{\lceil i \rceil}(i_2 p_2 x)
\end{eqnarray*}
Trivially, S2TIV(0) is true. For any $n \geq 0$, we now prove that if
S2TIV($n$) is true, then S2TIP($n+1$) and S2TIV($n+1$) are also true.
We first prove S2TIP($n+1$).   We calculate

~~~~$\pi^{[n+1]*1}_{\lceil i \rceil}(i_1 p_1 x)$ 

$=\argmax_{a} 
   \sum_{i'_1 p'_1 x'} P^1(i'_1 p'_1 x'|i_1 p_1 x,a)
   \left( i(x,x') +
   \gamma \; V^{[n]*1}_{\lceil i \rceil}((i'_1 p'_1 x') \right)$

~~~~
\begin{tabular}{l}
// For all $p'_1\neq i_1$ we have $P^1(i'_1 p'_1 x'|i_1 p_1 x,a)=0$, so
can simplify
\end{tabular}

$=\argmax_{a} 
   \sum_{i'_1 x'} P^1(i'_1 i_1 x'|i_1 p_1 x,a)
   \left( i(x,x') +
   \gamma \; V^{[n]*1}_{\lceil i \rceil}((i'_1 i_1 x') \right)$

~~~~
\begin{tabular}{l}
// Definition of $P^1$, $\text{S2TIV}(n)$ implies that we can replace
the term $V^{[n]*1}_{\lceil i \rceil}(i'_1 i_1 x')$ \\
// with $V^{[n]*2}_{\lceil i \rceil}(i_c p_c x')$, for
any constants $i_c$ and $p_c$, without changing the value
\end{tabular}

$=\argmax_{a} 
   \sum_{i'_1 x'}
   P^{I1}(i'_1|i_1 p_1 x,a)
   P^{X}(x'|i_1 p_1 x,a)
   \left( i(x,x') + \gamma \; V^{[n]*2}_{\lceil i \rceil}(i_c p_c x') \right)$

~~~~
\begin{tabular}{l}
// (C2)
\end{tabular}

$=\argmax_{a} 
   \sum_{i'_1 x'}
   P^{I1}(i'_1|i_1 p_1 x,a)
   P^{X}(x'|i_2 p_2 x,a)
   \left( i(x,x') + \gamma \; V^{[n]*2}_{\lceil i \rceil}(i_c p_c x') \right)$

$=\argmax_{a} 
   \sum_{x'}
   \left( \sum_{i'_1} P^{I1}(i'_1|i_1 p_1 x,a) \right)
   P^{X}(x'|i_2 p_2 x,a)
   \left( i(x,x') + \gamma \; V^{[n]*2}_{\lceil i \rceil}(i_c p_c x') \right)$

$=\argmax_{a} 
   \sum_{x'}
   \left( 1 \right)
   P^{X}(x'|i_2 p_2 x,a)
   \left( i(x,x') + \gamma \; V^{[n]*2}_{\lceil i \rceil}(i_c p_c x') \right)$

$=\argmax_{a} 
   \sum_{x'}
   \left( \sum_{i'_2} P^{I2}(i'_2|i_2 p_2 x,a) \right)
   P^{X}(x'|i_2 p_2 x,a)
   \left( i(x,x') + \gamma \; V^{[n]*2}_{\lceil i \rceil}(i_c p_c x') \right)$

$=\pi^{[n+1]*2}_{\lceil i \rceil}(i_2 p_2 x)$

The same calculation can be made for S2TIV($n+1$), by replacing the
$\argmax$ above with $\max$. With natural induction, this proves
S2TIP($n$) for all $n \geq 1$, which implies S2TI and, by S1, S2T.
\hfill $\Box$.
\\[.5ex]
{\bf Proof of S2.} In the same way as in the earlier proof of
S1 from S1T, S2 follows from S2T.  \hfill $\Box$.


\end{document}